\documentclass{article}
\usepackage{iclr2027_conference,times}
\usepackage{amsmath,amsfonts,bm}

\def\eqref#1{equation~\ref{#1}}
\def\1{\bm{1}}

\DeclareMathAlphabet{\mathsfit}{\encodingdefault}{\sfdefault}{m}{sl}
\SetMathAlphabet{\mathsfit}{bold}{\encodingdefault}{\sfdefault}{bx}{n}

\usepackage{iftex}
\ifXeTeX
  \usepackage{fontspec}
\fi
\usepackage{xcolor}
\definecolor{leverdata}{RGB}{185,30,30}
\definecolor{levermodel}{RGB}{25,70,190}

\usepackage{hyperref}
\usepackage{url}
\usepackage{booktabs}
\usepackage{array}
\usepackage{multirow}
\usepackage{graphicx}
\usepackage{amsmath,amssymb}
\usepackage{xcolor}
\usepackage{subcaption}
\usepackage{enumitem}
\usepackage{microtype}
\usepackage{pifont}
\usepackage{wrapfig}
\usepackage{caption}
\makeatletter
\renewcommand\paragraph{\@startsection{paragraph}{4}{\z@}{0.1ex plus 0.2ex minus 0.1ex}{-1em}{\normalsize\bf}}
\makeatother

\ifx\figurename\undefined \def\figurename{Figure}\fi

\newcommand{\Sect}[1]{Section~\ref{#1}}
\newcommand{\Fig}[1]{Figure~\ref{#1}}
\newcommand{\Tbl}[1]{Table~\ref{#1}}
\newcommand{\Equ}[1]{Eq.~\ref{#1}}

\newcommand{\Apx}[1]{Appendix~\ref{#1}}

\newcommand{\RNum}[1]{\uppercase\expandafter{\romannumeral #1\relax}}

\newcommand{\half}{$^{\scriptscriptstyle 1/2}$}

\newcommand{\Dist}{D}

\newcommand{\Nsteps}{N}

\title{Specialists Learn Faster:\\ Knowledge Organization in Pre-Training Shapes Knowledge Update Efficiency}

\author{Sheng Zhao \\
Department of Computer Science \\
University of Rochester \\
\texttt{szhao32@ur.rochester.edu}
\And
Yuhao Zhu \\
Department of Computer Science \\
University of Rochester \\
\texttt{yzhu@rochester.edu}
}

\iclrfinalcopy

\begin{document}

\maketitle
\lhead{Preprint}

\begin{abstract}
An ideal (language) model must continually update its knowledge.
Existing work largely focuses on making these updates efficient after training.
We investigate how the organization of time-varying knowledge during pre-training shapes the efficiency of subsequent updates.
Models trained on chronologically ordered knowledge specialize in the most recently encountered knowledge; we call them \textit{specialists}.
In contrast, models trained on the same knowledge in randomly shuffled order perform well across all represented time periods; we call them \textit{generalists}.
Surprisingly, specialists require far fewer update steps than generalists to acquire new knowledge.
To explain this advantage, we derive a law from first principles that relates update cost to the model's optimization dynamics and that closely predicts empirical results.
Guided by these findings, we design a strategy for organizing pre-training data that preserves stable historical knowledge while enabling efficient updates to current knowledge.
We also show that the benefits of temporal knowledge organization extend to real-world pre-trained models and existing knowledge update tools, and generalize to other knowledge orderings. 
Project page at: {\hypersetup{pdfborder={0 0 0}}\href{https://horizon-lab.org/specialist/}{\textcolor{blue}{https://horizon-lab.org/specialist/}}}.

\end{abstract}

\section{Introduction}
\label{sec:intro}

Both generalists and specialists have evolved in nature~\citep{sachdeva2020tuning}.
Generalists are fit across a range of environments although they are not particularly optimal for a particular environment.
Specialists are well adapted to a particular environment but lack fitness outside their niche.

We show that generalists and specialists can emerge in language models, influenced by how knowledge is ordered and presented during training (\Sect{sec:shape}).
Without losing generality, we focus on biographical knowledge, which has a natural order --- chronological order (\Sect{sec:setup}).
We find that models trained on chronologically ordered knowledge tend to become specialists: they perform best on the most recently encountered knowledge but poorly on knowledge from other periods.
In contrast, models trained on the same knowledge in randomly shuffled order tend to become generalists: they perform reasonably well across all periods represented in training.

Perhaps surprisingly, specialists acquire new knowledge more efficiently during post-training updates (\Sect{sec:update}).
To explain this advantage, we derive a law from first principles that describes knowledge-update dynamics in the model's optimization landscape.
The law relates update cost to two quantities: the pre-trained model's initial gap in acquiring the new knowledge (the knowledge gap) and how quickly the model closes that gap (the update speed).
Together with a gradient analysis, the law shows that specialists' advantage lies primarily in their larger initial gradient magnitudes, which allow them to close the (sometimes larger) knowledge gap much faster than generalists.

Our observations have practical values: an ideal (language) model must continually update its knowledge \citep{vu2024freshllms, johnson2025exploring} because, for instance, people change jobs over time or software APIs update across versions.
After deployment, a model is usually updated through model editing \citep{meng2022rome,meng2023memit,decao2021editing,yao2023editing} or continued pre-training \citep{gururangan2020dont,ibrahim2024simple}.
Most prior work focuses on making these updates cheaper and less harmful \textit{once the model has already been trained} by, for example, meta-learning an editor \citep{mitchell2022fast}, storing updates in external memory \citep{mitchell2022memory,hartvigsen2023grace}, or constraining updates to spare other knowledge \citep{fang2025alphaedit}.

These post-training approaches apply after a model has learned its knowledge.
Our work studies the role of the original training data in update efficiency.
Building on our findings, we design a strategy for organizing pre-training data that preserves rarely changing historical knowledge while enabling efficient updates to recent knowledge (\Sect{sec:frontier}).
We also show that the benefits of temporal knowledge organization extends to real-world pre-trained models (\Sect{sec:cpt}) and existing knowledge update tools (\Sect{sec:plastic-edit}), and generalize to other knowledge orderings (\Sect{sec:apply}).

In summary, we make three main contributions.

\begin{itemize}[leftmargin=1.2em, labelsep=0.4em, topsep=0pt, partopsep=0pt, itemsep=1pt, parsep=0pt]

\item We show that models trained on chronologically ordered knowledge require roughly half as many steps to learn new knowledge as those trained on temporally shuffled knowledge.

\item We derive a law governing the update cost.
The law shows that the temporal order of knowledge reduces update cost primarily by substantially increasing the initial gradient of the model.

\item Using our findings, we show how to design temporal knowledge organizations to train models that are both generalists and specialists.
The benefits extend to pre-trained (real-world) models and existing knowledge update tools, and generalize to other knowledge orderings.

\end{itemize}

\section{Related Work}

\paragraph{Time-Varying Knowledge.}
Temporal benchmarks \citep{lazaridou2021mind,jang2022temporalwiki,jang2022towards,loureiro2022timelms,vu2024freshllms}, studies of outdated knowledge \citep{zhao2024set}, date conditioning \citep{dhingra2022time} and probes of temporal representations \citep{gurnee2024language,park2025temporal} ask what a model knows about the past and the present.
We instead ask how the history of a changing fact is represented and what it costs to change the fact again.

\begin{wraptable}{r}{0.56\linewidth}
\vspace{-10pt}
\scriptsize
\setlength{\tabcolsep}{4pt}
\renewcommand{\arraystretch}{1.15}
\newcommand{\lvd}[1]{\textcolor{leverdata}{#1}}
\newcommand{\lvm}[1]{\textcolor{levermodel}{#1}}
\caption{Prior work by stage and task. \lvd{Red}: through data organization and training dynamics; \lvm{Blue}: through architecture or hyperparameters. Ours is the only data organization-level work on efficient editing at pre-training stage.
}
\label{tab:axes}
\begin{tabular}{@{}>{\raggedright\arraybackslash}p{1.25cm}>{\raggedright\arraybackslash}p{3.0cm}>{\raggedright\arraybackslash}p{3.0cm}@{}}
\toprule
 & Efficient updating \& editing & Catastrophic forgetting \\
\midrule
Pre-training & {\lvd{\textbf{Ours}}}\newline \lvm{\citet{liu2025drop}, \citet{kim2025knowledge}} & \lvd{\citet{toneva2019empirical}, \citet{tirumala2022memorization}, \citet{ibrahim2024simple}}\newline \lvm{\citet{ke2023continual}, \citet{dohare2024loss}} \\
\addlinespace
Post-training & \lvd{\citet{ovadia2024fine}, EntiGraph \citep{yang2025synthetic}, SEAL \citep{zweiger2025self}}\newline \lvm{ROME \citep{meng2022rome}, MEMIT \citep{meng2023memit}, MEND \citep{mitchell2022fast}} & \lvd{\citet{scialom2022fine}, \citet{luo2023empirical}, \citet{gekhman2024does}}\newline \lvm{EWC \citep{kirkpatrick2017overcoming}, GRACE \citep{hartvigsen2023grace}, WISE \citep{wang2024wise}} \\
\bottomrule
\end{tabular}
\vspace{-10pt}
\end{wraptable}

\paragraph{Temporal Order in Training.}
Data order has been studied as curriculum learning \citep{bengio2009curriculum,wu2021when} and, for growing corpora, as chronological or continual pre-training \citep{jin2022lifelong,li2025tic, pilchen2026understanding}.
We hold the facts fixed and vary only their order, which isolates its effect on representation and update cost.

\paragraph{Knowledge Update.}
Prior work on knowledge updating differs in when it intervenes and what it changes (see \Tbl{tab:axes}).
After pre-training, editors rewrite facts in place \citep{meng2022rome,meng2023memit,fang2025alphaedit}, learn an editor \citep{mitchell2022fast,zweiger2025self} or modify the update data \citep{ovadia2024fine,yang2025synthetic}.
We ask whether data organization eases later updates.

\section{Experimental Setup}
\label{sec:setup}

\paragraph{Temporal biographies.}
Our dataset extends the synthetic biographies of \citet{allenzhu2023physics} with a time axis.
It generates 25{,}000 individuals (at 110M; more at larger scales), each with a unique name and six attributes: birthday, birth city, university, field of study, employer and employer city; each attribute has a corresponding value.
Each individual appears 1{,}000 times between 2000 and 2024. 
In each year, individuals hold a value for each attribute, and some of these values change over time:
their birthday and birth city remain unchanged, university and field change a few times, and their employer and employer's city change more frequently.
Each sentence is drawn from about fifty templates for its attribute.
An example is shown below (see \Apx{app:protocol} for more details):
\par\begingroup\small\addtolength{\leftskip}{1.2em}\addtolength{\rightskip}{1.2em}\vspace{3pt}\noindent
In 2010, John Ivy Bi was born on \underline{December 17, 1979} in \underline{Moreno Valley, CA}. He studied at \underline{Bowdoin College} and majored in \underline{Photography}. He worked at \underline{Under Armour} in \underline{San Jose, CA}.
\par\vspace{-3pt}\endgroup\noindent
In this example, in year $Y = 2010$, the person ``John Ivy Bi'' and the attribute ``birthday'' form a fact identity $I$, which has a value $V =$ ``December 17, 1979'';
there are six such identity-value pairs for each person in each year (see \Fig{fig:temporal-example} for more details).

\paragraph{Pre-training organization and models.}
From the same records we build six pre-training corpora by varying two factors: record order and timestamp inclusion.
A chronological order (C) presents each person's records in the chronological order, sorted by the timestamps, which are the ``In \{year\}, '' prefix of the example above.
Records of different people are unordered.
A shuffled order (S) shuffles all records in an random order.
For each order, we build three variants, one that does not carry the timestamp (i.e., no ``In \{year\}, '' prefix), one carries the timestamp (D), and a third that carries timestamp on randomly chosen half of the records (D$^{1/2}$).
This yields six training corpora (C, S, CD, SD, CD$^{1/2}$, and SD$^{1/2}$).
We pre-train GPT-2 style decoders with rotary position embeddings and the GPT-2 tokenizer at three scales; 110M, 350M, and 682M (see \Tbl{tab:arch} for model details).
The pre-training loss is the standard cross-entropy over the whole record.

\begin{wrapfigure}{r}{0.44\linewidth}
\vspace{-12pt}
\centering
\includegraphics[width=\linewidth]{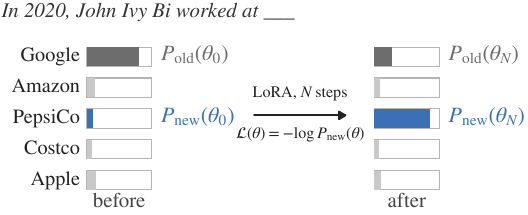}
\caption{Example of an \textit{overwrite} update.
Bars represent the model's predicted probabilities over an attribute's values.
Starting from initial parameters $\theta_0$, a LoRA adapter is trained for $N$ steps on the loss $\mathcal{L}(\theta)=-\log P_{\mathrm{new}}(\theta)$, until the new value's probability passes its competitor's, $P_{\mathrm{old}}$ (the strongest value model held before updating).}
\label{fig:edit}
\vspace{-8pt}
\end{wrapfigure}

\paragraph{Updating.}
After pre-training, we consider two types of knowledge updates: 
an \textit{overwrite} update corrects an existing record, while an \textit{injection} update adds a record for a new person.
Each update is performed by training a LoRA adapter~\citep{hu2022lora}.
\Fig{fig:edit} shows an overwrite update: the company John Ivy Bi worked at in 2020 is corrected from Google to PepsiCo.
This is done by training the adapter on ``In 2020, John Ivy Bi worked at PepsiCo.''
Before LoRA, $P_{\mathrm{old}}(\theta_0)$, the probability of the old value Google, is the highest, whereas the probability of the new/updated value PepsiCo, $P_{\mathrm{new}}(\theta_0)$, is very low.
After $N$ LoRA steps when model parameters are updated from $\theta_0$ to $\theta_N$, $P_{\mathrm{new}}(\theta_N)$ becomes the highest.
The fine-tuning uses the cross-entropy loss computed only on the new value, i.e., $-\log P_{new}(\theta)$, and stops when $P_{\mathrm{new}}(\theta)$ passes the competitor's probability $P_{\mathrm{old}}(\theta)$.
\footnote{
We mark an update successful once $P_{\mathrm{new}}$ exceeds the competitor, similar to prior model-editing metrics \citep{meng2022rome}; \Tbl{tab:strict} confirms the same conclusions under stricter criteria empirically.
}

\paragraph{Knowledge Query.}
A query asks for the value of one attribute of a person.
We experiment with two forms of knowledge querying.
A \emph{dated query} specifies a year along with a person's name and an attribute and counts the response as correct if the answer is any value held during that year (since a value can change within a year).
An \emph{undated query} omits the year and expects the latest value.

\section{The Emergence of Generalists and Specialists}
\label{sec:shape}

This section studies how the temporal organization of knowledge exposed during training shapes what a model learns.
We extend the results to other ordering in \Sect{sec:apply}.

\subsection{Organizing Time-Varying Knowledge Influences Generalists vs. Specialists}
\label{sec:order}

\begin{figure}[t]
\centering
\includegraphics[width=0.95\linewidth]{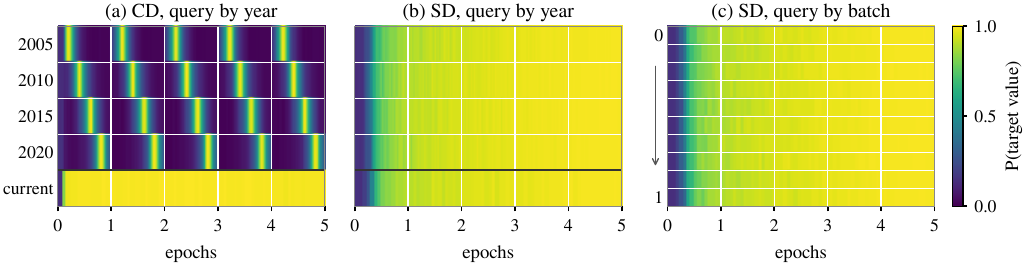}
\vspace{-3pt}
\caption{
Query accuracy of the 110M CD and SD models over five training epochs; color represents accuracy.
(a, b) Query by year (for CD and SD).
Rows 2005--2020 query knowledge of a year (e.g., ``In 2005, \ldots''), and the ``current'' row queries knowledge from the most recent batch the model is trained on.
(c) Query by batch (for SD).
Each row queries knowledge from one of the ten batches in the training corpus, from the first batch (0) to the last (1).
See \Apx{app:protocol} for details.
}
\vspace{-3pt}
\label{fig:tracking}
\end{figure}

To demonstrate how the organization of time-varying knowledge during training can influence the model, \Fig{fig:tracking} compares the training dynamics across three settings:
\begin{itemize}[leftmargin=1.2em, labelsep=0.4em, topsep=0pt, partopsep=0pt, itemsep=1pt, parsep=0pt]
    \item On the left, the model is trained on the CD corpus, and constantly being queried at different batches and epochs during training ($x$-axis) using four dated queries ($y$-axis) --- Year 2005, 2010, 2015, and 2020 --- as well as the ``current'' query, which queries using the knowledge trained in the immediately previous batch.
    The heatmap represents the query accuracy, averaged over all the queries in the corresponding query category.
    \item The middle figure shows the same training dynamics but using the SD corpus for training.
    \item Finally, the last figure shows the dynamics when the model is trained on the SD corpus, but is queried using different batches of the training corpus.
    Importantly, the querying batch order ($y$-axis) matches that of training.
\end{itemize}

Comparing the first two figures, we see that the CD-trained model tends to learn only the knowledge that it is just exposed to, and quickly forgets the knowledge in the past.
As a result, there are distinct phases during training where each dated query category has a high accuracy; this pattern repeats at every epoch.
The model essentially transitions across different specialists, each specializing to knowledge in a given time span.
In contrast, the model trained using the SD corpus tends to yield generalists, which perform well on all dated queries \textit{even in the first epoch} when the model has not experienced all the training knowledge yet.

\begin{figure}[t]
\centering
\includegraphics[width=0.92\linewidth]{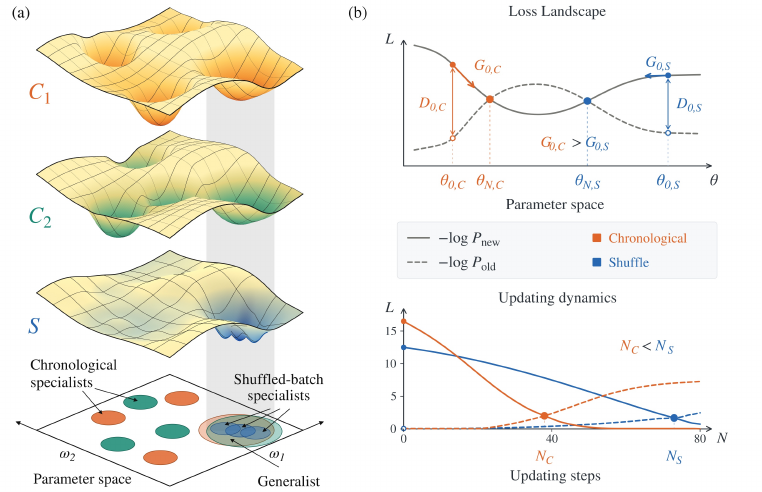}
\caption{Intuition for generalists vs. specialists and for why specialists update knowledge faster.
(a) Loss landscapes of knowledge in two years ($C_1$ and $C_2$) and of all knowledge ($S$) over the same model parameter space (bottom plane).
Each valley in $C_1$ and $C_2$ is a chronological specialist that learns the knowledge of that year well.
Every valley in $S$ is a shuffled-batch specialist for a batch during shuffled training.
The specialists of $C_1$ (orange) and $C_2$ (green) overlap in a region, which is where generalists lie.
Different shuffled-batch specialists (blue) lie close to each other and all fall inside the generalist region.
(b) Top: the loss of the new value ($-\log P_{\mathrm{new}}$, solid) and the old value ($-\log P_{\mathrm{old}}$, dashed) across the parameter space $\theta$.
An update starts at $\theta_{0, C}$ and $\theta_{0, S}$ for a specialist and a generalist, respectively; $D_{0, C}$ and $D_{0, S}$ are the initial learning gap, respectively, and $G_{0, C}$ and $G_{0, S}$ are the corresponding initial gradients.
$\theta_{N, C}$ and $\theta_{N, S}$ are the points where the two losses cross, i.e., where each update completes, in the two cases, respectively.
Bottom: the actual change of loss over update steps, comparing a 110M specialist and a 110M generalist; an update ends where they cross ($P_{\mathrm{new}}=P_{\mathrm{old}}$), and the specialist crosses first ($N_C<N_S$).}
\label{fig:landscape}
\end{figure}

\Fig{fig:landscape} (a) shows intuitively how shuffled training yields generalists.
Each year (two here, $C_1$ and $C_2$) has an optimization landscape with multiple specialists, each of which learns the knowledge in that year well and corresponds to a valley.
Importantly, the specialists of different years could overlap.
The overlapped region contains the ``jacks-of-all-trades'': they are generalists that are well-optimized for knowledge across years.
Given that specialists generally far outweigh generalists in number, a random initialization will much more likely end up in a specialist under chronological training.

Why do models trained under the shuffled corpora end up in the generalist region?
When trained under SD, each batch yields a specialist model of that particular batch, i.e., the ``shuffled-batch specialists'' in \Fig{fig:landscape} (a).
Different shuffled batches can be thought of as samples of the same underlying corpus;
as a result, different batch's specialists are very close to each other in the landscape and all fall in the generalists region.
This intuition is confirmed in \Fig{fig:tracking} (c).
Apart from the initial stage in the first epoch, the SD-trained model performs well for all batches, even for unseen batches (in the first epoch).
This suggests that shuffled-batch specialists are similar generalists.

\subsection{The Role of Time in Chronological Training}
\label{sec:time}

A key assumption in \Fig{fig:landscape} (a) is that specialists and generalists, trained under CD, perform equally well on the year for which the specialist is optimized.
Yet, a specialist clearly has not learned the year information, as shown by its poor performance under dated queries of other years (see \Fig{fig:tracking} (a)).
Why, then, can a model that has not learned the year information perform as well as one that has?

Intuitively, this is because under chronological training, nearby records of the person usually share the same value, because a person's attributes (e.g., job) do not change rapidly in a short period of time; 
\Apx{app:protocol} provides detailed statistics on this.
As a result, the loss can be reduced without relying strongly on time and, accordingly, the gradients would unlikely point toward regions in the landscape that relies on time.
Under shuffling, however, different values of the same fact can appear close together, so the model must use the year to tell them apart.

To formalize this, we represent each fact by its identity $I$, year $Y$, and value $V$ (see \Sect{sec:setup} for notation convention and \Fig{fig:temporal-example} for an annotated example).
For an empirical distribution $Q_{t,W}$ over $W$ consecutive fact records starting at $t$, the value-prediction part of the training loss is:
\begin{equation}
L_{t,W}(p)=\mathbb{E}_{(I,Y,V)\sim Q_{t,W}}[-\log p(V\mid I,Y)].
\label{eq:temporal-objective}
\end{equation}
We quantify the predictive benefit of the year by comparing a year-independent predictor $r(V\mid I)$ with a year-conditioned predictor $p(V\mid I,Y)$. The difference between their minimum losses is
\begin{equation}
R_{t,W}:=\underbrace{\inf_r L_{t,W}(r)}_{\text{without year}}-\underbrace{\inf_p L_{t,W}(p)}_{\text{with year}}=\operatorname{MI}_{Q_{t,W}}(V;Y\mid I),
\label{eq:temporal-risk}
\end{equation}

where $\inf(\cdot)$ is the infimum operator,  $R_{t,W}$ is how much the loss would reduce with the additional year $Y$ information, and $\operatorname{MI}$ denotes conditional mutual information, representing how much the year $Y$ tells us about the value $V$ once the fact identity $I$ is known.
For the detailed derivation of Equations~\ref{eq:temporal-objective}--\ref{eq:temporal-risk}, see \Apx{app:temporal-risk}.

Empirically, $R_{t,W}$ is close to 0 under CD training (see \Apx{app:temporal-risk} for details).
Intuitively, this means that predicting $V$ does not depend much on knowing the year information $Y$ when the person's identity $I$ is known.
Therefore, year-agnostic specialists and year-aware generalist have similar query performance.
\Apx{app:time-patching} further shows that the year information in a specialist model is \textit{not} simply dropped; it does participate in model inference, but the information is treated as an ordinary token and does not contribute much to the prediction.

\section{Efficiency of Updating Knowledge}
\label{sec:update}

With generalists and specialists naturally emerging from training under different temporal organizations of knowledge, we now turn our attention to how easy it is for them to learn new knowledge.

\subsection{Specialists Acquire New Knowledge Faster}
\label{sec:cost}

How the knowledge is exposed to a model at the pre-training time affects how the model acquires new knowledge later.
\Tbl{tab:steps} compares the number of updates required by specialists and that by generalists across three model sizes and six training corpora.
Specialists require 27\%-58\% fewer steps to both overwrite a fact or inject a new fact than generalists (\Apx{app:tiers} for more results).
The saving increases with the model size, suggesting that more powerful generalists are even harder to update knowledge for.

These trends also hold when the training corpus omits timestamps (C vs. S) or includes them in only half of the records (CD$^{1/2}$ vs. SD$^{1/2}$).
Updates are undated for C and S, and evenly split between dated and undated for CD$^{1/2}$ and SD$^{1/2}$.
\Apx{app:robust} shows that the same pattern holds across optimizers and learning rates, and when full fine-tuning replaces LoRA.

\begin{figure}[t]
\centering
\includegraphics[width=.9\linewidth]{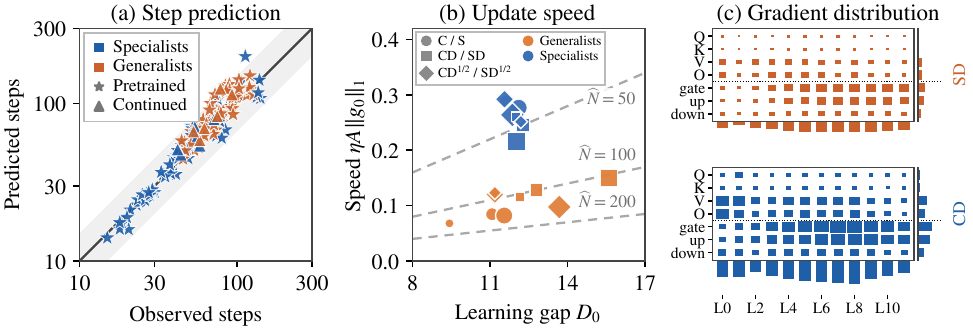}
\vspace{-3pt}
\caption{Update cost under chronologically ordered and shuffled training.
(a) Predicted updating steps against observed steps;
Each of the 18 models (3 model sizes in \Tbl{tab:arch} $\times$ 6 training corpora) is shown in two variants: pre-trained from scratch (stars) and continued from a pre-trained model (triangles).
(b) We plot these 18 pre-trained models using their initial distance $D_0$ and the learning speed $\eta A\|g_0\|_1$; each point is a median over 160 updates, and the dashed lines mark a constant number of steps (\Equ{eq:steplaw}).
Marker sizes are proportional to the model sizes.
(c) Contribution to the initial gradient $\|g_0\|_1$ from each layer and each block within a layer, for a generalist (above; trained on SD) and a specialist CD (below; trained on CD);
the area of a square is the contribution of that block, the histograms sum the contributions by layer and by block (see details in \Apx{app:derivation}).
}
\label{fig:cost}
\end{figure}

\begin{table}[t]
\centering
\footnotesize
\caption{Update cost comparison between different model sizes and training strategies.
We use three model sizes, and each model is trained under the six training corpora.
An \textit{overwrite} update changes a known person's latest value and an \textit{inject} update adds a fact about an unseen person. 
The ``saved'' column is the percentage of update steps that a chronological model saves over the corresponding shuffled model (all results have a paired Wilcoxon $p<0.05$).
We also show the $D_0$ ratio and the $\|g_0\|_1$ ratio between a corresponding specialist-generalist pair.
}
\label{tab:steps}
\scriptsize                       
\setlength{\tabcolsep}{4.5pt}
\renewcommand{\arraystretch}{1.0}
\begin{tabular}{llcccccccc}
\toprule
 & & \multicolumn{4}{c}{Overwrite} & \multicolumn{4}{c}{Inject} \\
\cmidrule(lr){3-6}\cmidrule(lr){7-10}
Scale & Pair & steps & saved & $D_0$ ratio & $\|g_0\|_1$ ratio & steps & saved & $D_0$ ratio & $\|g_0\|_1$ ratio \\
\midrule
\multirow{3}{*}{110M} & C / S & 51.0 / 88.0 & \textbf{$+$42\%} & 1.27$\times$ & 2.82$\times$ & 37.0 / 59.0 & \textbf{$+$37\%} & 1.18$\times$ & 2.68$\times$ \\
 & CD / SD & 53.0 / 73.0 & \textbf{$+$27\%} & 1.03$\times$ & 2.02$\times$ & 37.0 / 52.0 & \textbf{$+$29\%} & 1.05$\times$ & 1.90$\times$ \\
 & CD\half{} / SD\half{} & 53.0 / 75.0 & \textbf{$+$29\%} & 1.00$\times$ & 1.61$\times$ & 33.0 / 55.0 & \textbf{$+$40\%} & 0.96$\times$ & 2.02$\times$ \\
\midrule
\multirow{3}{*}{350M} & C / S & 50.0 / 116 & \textbf{$+$57\%} & 1.00$\times$ & 2.71$\times$ & 31.5 / 65.0 & \textbf{$+$52\%} & 0.77$\times$ & 2.44$\times$ \\
 & CD / SD & 55.0 / 90.0 & \textbf{$+$39\%} & 0.83$\times$ & 1.63$\times$ & 32.0 / 57.0 & \textbf{$+$44\%} & 0.79$\times$ & 1.45$\times$ \\
 & CD\half{} / SD\half{} & 50.5 / 94.0 & \textbf{$+$46\%} & 0.87$\times$ & 1.76$\times$ & 31.5 / 61.0 & \textbf{$+$48\%} & 0.80$\times$ & 1.72$\times$ \\
\midrule
\multirow{3}{*}{682M} & C / S & 49.0 / 118 & \textbf{$+$58\%} & 1.10$\times$ & 2.40$\times$ & 29.0 / 63.5 & \textbf{$+$54\%} & 0.68$\times$ & 2.11$\times$ \\
 & CD / SD & 54.0 / 87.5 & \textbf{$+$38\%} & 0.72$\times$ & 1.39$\times$ & 31.0 / 54.5 & \textbf{$+$43\%} & 0.59$\times$ & 1.22$\times$ \\
 & CD\half{} / SD\half{} & 50.0 / 100 & \textbf{$+$50\%} & 0.85$\times$ & 1.91$\times$ & 29.0 / 60.0 & \textbf{$+$52\%} & 0.64$\times$ & 1.60$\times$ \\
\bottomrule
\end{tabular}
\end{table}

\subsection{A Step Law for Updating Efficiency}
\label{sec:derive}

\providecommand{\qold}{q_{\mathrm{old}}}
\providecommand{\gz}{g_0}
\providecommand{\Pnew}{P_{\mathrm{new}}}
\providecommand{\Pold}{P_{\mathrm{old}}}

We now aim to understand what determines the update cost.
To that end, we derive an empirical law by viewing learning an update as a trajectory through the optimization landscape.
Recall from \Sect{sec:setup} that an update, be it an overwrite or an injection, stops at step $N$ when $D_N = \log P_{old}(\theta_N) - \log P_{new}(\theta_N) < 0$.
Therefore, $N$ depends on two factors: the initial distance $D_0$ in the pre-trained model (i.e., the knowledge gap) and how fast the distance is closed per fine-tuning step (i.e., the update speed).

We omit the derivation and refer readers to \Apx{app:derivation}.
In short, the relationship is given by:
\begin{equation}
\widehat{\Nsteps}
=
\frac{\Dist}{\eta A \|\gz\|_1},
\label{eq:steplaw}
\vspace{-4pt}
\end{equation}

where $\eta$ is the learning rate in the optimizer (Adam), $\gz=\nabla_\theta[-\log\Pnew](\theta_0)$ is the initial gradient before the first fine-tuning step and represents the initial learning speed;
$A$ is a single calibration constant that is applied across all training settings.
$A$ is calibrated such that $A\|\gz\|_1$ represents the \textit{average} update speed, i.e., how much each fine-tuning step closes the initial gap $D_0$.

The derivation of \Equ{eq:steplaw} does not depend on whether, and how often, timestamps are part of the training knowledge.
Therefore, the law applies to all 18 variants in \Tbl{tab:steps}.
\Fig{fig:cost}(a) compares the observed $N$ and the predicted $N$ using \Equ{eq:steplaw}.
The model accurately captures the update cost across model scales, timestamp settings, and training organizations; the overall Pearson correlation coefficient $r = 0.91$.
Without refitting $A$, the same model also predicts update cost after continued pre-training changes the model's organization (\Apx{app:cpt}), and predicts well across other learning rates and with another optimizer (\Apx{app:robust}).

Since $A$ and $\eta$ are constant in \Equ{eq:steplaw}, differences in update cost between generalists and specialists arise from $D_0$ and $\gz$.
\Fig{fig:cost} (b) plots each pre-trained model by its $D_0$ ($x$-axis) and $A\|\gz\|_1$ ($y$-axis).
The three lines correspond to three different update-step counts.
We make two main observations.

First, update speed contributes more than the initial knowledge gap to specialists' update advantage.
\Fig{fig:cost} (b) shows that generalists have a wide range of the initial learning gap $D_0$ than that of the specialists.
Regardless of the initial learning gap, however, specialists consistently have a much higher speed of learning than that of generalists (2.5 $\times$ higher on average).
As an example, even though the 110M specialist trained under CD has actually a slightly (1.03 $\times$) larger knowledge gap than the generalist trained under SD, the former updates more than 2 $\times$ faster and ends up acquiring the knowledge much sooner.
Such dynamics are illustrated in \Fig{fig:landscape} (b).

Second, \Fig{fig:cost} (b) also shows that larger models tend to have larger initial learning gaps (model size is proportional to the marker sizes).
This is because larger generalists are trained better, so they are naturally more confident in the old knowledge before the update, leaving a larger knowledge gap to close during update.
This explains the greater savings observed for larger models in \Tbl{tab:steps}.

\Fig{fig:cost} (c) further visualizes the contribution to the initial gradient from each layer and each block within a layer between a specialist and a generalist.
We see that the absolute amplitude of initial gradient increases significantly from generalists to specialists.
In particular, the middle layers contribute the most to the specialists' larger gradient, which is where the faster update comes from.

\section{Applications of Ordering Knowledge in Pre-Training}
\label{sec:mix}

Our analyses so far demonstrate a trade-off between generalists and specialists --- influenced by the temporal ordering of knowledge.
This section shows four applications of this observation to improve language models in practice.

\subsection{Mixing Generalists and Specialists}
\label{sec:frontier}

\paragraph{Setup.}
Consider a deployed model whose more recent knowledge changes frequently, while its historical knowledge remains fixed.
Ideally, the pre-trained model should preserve its historical knowledge accurately while efficiently acquiring new knowledge.

Our idea is to train a generalist for the history and a specialist for the present.
Specifically, we select a recent time window near the present and organize each person's record within this window chronologically, while shuffling all records from the earlier history beyond the window.
This forms the corpus Mix-C.
As a comparison, Mix-S contains the same knowledge as Mix-C but shuffles the records randomly.
Both models are trained at the 110M scale; full construction and training details are provided in \Apx{app:practical-frontier}.

\begin{figure}[t]
\centering
\includegraphics[width=0.95\linewidth]{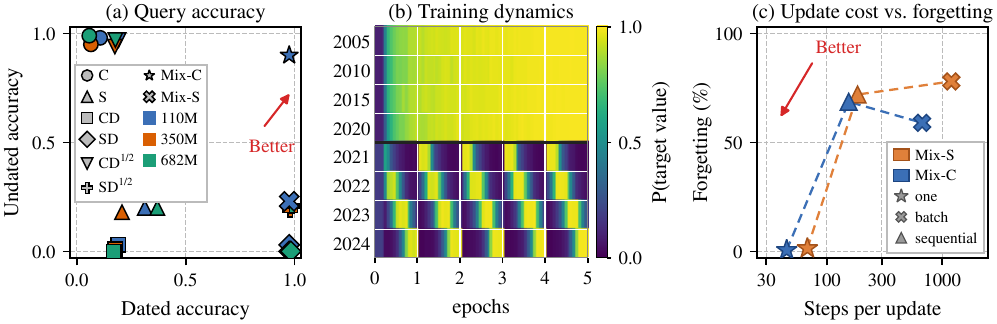}
\vspace{-3pt}
\caption{Query accuracy and update cost of the Mix corpora (\Sect{sec:frontier}). (a) Dated query accuracy (2005--2020 average) against undated query accuracy of different pre-trained models.
(b) Dated query accuracy of Mix-C through five epochs of training, for historical years (2005--2020) whose knowledge is shuffled and for recent records (2021--2024) whose knowledge is ordered.
(c) Median steps per update (log scale) against forgetting (percentage of other people's facts that are incorrectly changed), for single updates (star), batch updates (cross) and sequential updates (triangle).}
\vspace{-5pt}
\label{fig:frontier}
\end{figure}

\paragraph{Performance.}
We find that Mix-C is able to preserve historical knowledge while acquiring new knowledge efficiently.
\Fig{fig:frontier} (a) show the knowledge query, both dated and undated, accuracy of Mix-C.
Among all training organizations, Mix-C is the only model that achieves high query accuracy on both historical knowledge (with dated queries) and current knowledge (with undated queries).
\Fig{fig:frontier} (b) shows the training dynamics of Mix-C.
The model behaves as a generalist on the historical knowledge, while exhibiting specialist-like dynamics on recent knowledge, consistent with the observations in \Sect{sec:shape}.

At the same time, the specialist-like organization of recent knowledge also enables faster acquisition of new knowledge.
\Fig{fig:frontier} (c) compares the \emph{overwrite} update cost of Mix-C and Mix-S under single, batch, and sequential updates;
we also show the percentage of knowledge forgetting on the $y$-axis.
A single update changes one fact, a batch update changes multiple facts jointly, and sequential updates apply a series of batches.
All updates are made to the most recent year; see \Apx{app:practical-frontier} for details.
Forgetting is the percentage of facts about other people whose answers are incorrectly changed by an update.
Mix-C consistently requires 16\%-44\% fewer update steps than Mix-S across all three settings, while having 5\%-57\% less forgetting on the historical records.

\subsection{Reshaping a Real-World Pre-trained Model}
\label{sec:cpt}

\paragraph{Setup.}
Pre-training from scratch is computationally expensive.
We therefore ask whether a short stage of continual pre-training can improve a model's update efficiency while preserving its historical knowledge.
We start from Gemma-2-9B-it \citep{team2024gemma}, an instruction-tuned model that was trained on real-world, general data including time-varying knowledge.

For continual pre-training, we sample historical records with timestamps and keep them temporally shuffled, while sampling a long recent time window and arranging each person's records chronologically.
We mix the two parts while preserving the within-person chronological order of the records in the window, forming the corpus C-C.
As comparisons, C-S contains exactly the same records but is randomly shuffled, and we also include the original pre-trained model (Base) as a baseline.
Further details are provided in \Apx{app:realdata}, and \Apx{app:cpt} presents results on our synthetic data.

\paragraph{Performance.}

\begin{figure}[t]
\centering
\includegraphics[width=.85\linewidth]{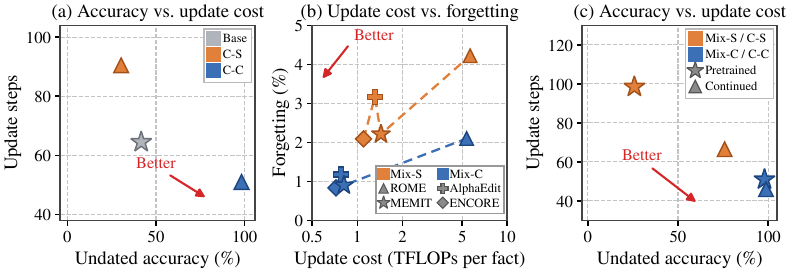}
\vspace{-3pt}
\caption{
(a) Undated query accuracy of the continual pre-training stage vs. the median update steps on \emph{overwrite} updates, for Gemma-2-9B-it (Base) and its two continually pre-trained variants.
(b) Update cost vs. percentage of forgetting for the four knowledge editors applied to the 110M Mix-C and Mix-S models; each model's four editors are joined by a dashed line.
(c) Undated query accuracy against the median update steps in the versioned-software dataset; results are for a 110M model.
}
\vspace{-3pt}
\label{fig:cases}\label{fig:reshape}\label{fig:editors}\label{fig:toolworld}
\end{figure}

\Fig{fig:cases} (a) shows that after continual pre-training using C-C, the model improves its undated query accuracy of current knowledge from 42\% to 98\%, while requiring 1.3$\times$ fewer steps than Base and 1.8$\times$ fewer steps than C-S during \emph{overwrite} updates.
Meanwhile, its general ability is preserved; \Apx{app:realdata} reports the forgetting results.

\subsection{Improving Published-Knowledge Editors}
\label{sec:plastic-edit}

\paragraph{Setup.}
So far, we have updated knowledge using LoRA.
Another common way of knowledge update is through dedicated knowledge-editing methods.
We next ask whether these methods could benefit from chronological pre-training.
We run ROME \citep{meng2022rome}, MEMIT \citep{meng2023memit}, AlphaEdit \citep{fang2025alphaedit} and ENCORE \citep{gupta2025encore}, on Mix-C and Mix-S, applying the \emph{overwrite} updates as in \Sect{sec:frontier}.
\Apx{app:practical-editing} contains more experimental details.

\paragraph{Performance.}
We find that specialists induced by temporal knowledge organization can further improve existing knowledge-editing methods over their shuffled counterparts.
\Fig{fig:cases} (b) compares the update cost (TFLOPs per update) and forgetting of these editors when applied to Mix-C, which contains a specialist for present knowledge (\Sect{sec:frontier}), and to its fully shuffled counterpart Mix-S.
For the same editing method, the specialist organization in Mix-C provides an additional reduction in both update cost and forgetting over Mix-S.
The gain reaches up to $46\%$ lower update cost with MEMIT \citep{meng2023memit} and $63\%$ less forgetting with AlphaEdit \citep{fang2025alphaedit}.

\subsection{Generalizing to Another Knowledge Ordering: Versioned Software Change}
\label{sec:apply}

\paragraph{Setup.}
So far, biographical knowledge is ordered chronologically, but do the same observations hold under other domains and orderings?
We study a practical case: software tools, where the version of a tool serves as the ordering key. 
For example, when an API changes its URL across versions, a model should quickly learn the API's latest endpoint without losing its knowledge of how to call other tools.
We build the tool-call dataset in a similar way to the biography dataset in \Sect{sec:setup}.
The dataset contains multiple tools, each described by six fields (for example, host, authentication header, etc.), each of which has a value.
An example is (see \Apx{app:toolworld} for more examples):
\par\begingroup\small\addtolength{\leftskip}{1.2em}\addtolength{\rightskip}{1.2em}\noindent
For v8, the Flint Payments API carries host \underline{api.flintpay.com}. 
Its input parameter is \underline{refund\_id}.
\par\endgroup\noindent
We follow the pre-training and continual pre-training setups in \Sect{sec:frontier}.
Specifically, for pre-training from scratch, we shuffle the older versions and present the recent versions in version order, as in Mix-C.
For continual pre-training, we start from a shuffled base model, present records within recent version window in version order, and replay its older versions in shuffled order, as in C-C.
We ask whether the same setup can learn a specialist for recent versions while keeping a generalist over older versions. 
\Apx{app:toolworld} shows more details about the training setup.

\paragraph{Performance.}
The same pattern appears in the tool-use domain.
\Fig{fig:cases} (c) shows that both Mix-C and C-C answer $98$--$99\%$ of queries about the latest version, while cutting update cost by up to $48\%$ and $33\%$ relative to Mix-S and C-S, respectively.
Both also keep their knowledge of older versions.

\section{Concluding Remarks}

Pre-training knowledge ordering shapes the cost of future knowledge updates.
Models trained on temporally ordered knowledge, which we call specialists, require roughly half as many update steps as generalists, models trained on shuffled knowledge.
We derive a step law to explain this advantage.
We use this insight to design new knowledge organization to preserve historical knowledge while enabling efficient updates.
The benefit of pre-training knowledge organization extends to already-trained models and existing post-training update tools.

Apart from time, \Sect{sec:apply} explores another ordering, where knowledge about software documentation is ordered by software version.
Regardless of the specific order $Y$, what matters is the additional information that $Y$ provides to a knowledge query given all other available information (see the mutual information \Equ{eq:temporal-risk}).
An interesting future investigation is to quantify how this mutual information influences the emergence and behaviors of generalists and specialists.

While our work focuses on language models, evolutionary biologists have long studied how generalists and specialists can evolve in time-varying environments~\citep{kassen2002experimental}.
Many also have looked at various advantage of specialists such as faster decision making~\citep{bernays1999specialists}.
Another interesting path is to investigate whether the learning benefit of specialists extend to evolution, which obtains knowledge through gradient-free methods.

\bibliography{refs}

@article{kassen2002experimental,
  title={The experimental evolution of specialists, generalists, and the maintenance of diversity},
  author={Kassen, Rees},
  journal={Journal of evolutionary biology},
  volume={15},
  number={2},
  pages={173--190},
  year={2002},
  publisher={Wiley Online Library}
}

@article{bernays1999specialists,
  title={Specialists make faster decisions than generalists: experiments with aphids},
  author={Bernays, Elizabeth A and Funk, Daniel J},
  journal={Proceedings of the Royal Society B: Biological Sciences},
  volume={266},
  number={1415},
  pages={151},
  year={1999}
}

@article{sachdeva2020tuning,
  title={Tuning environmental timescales to evolve and maintain generalists},
  author={Sachdeva, Vedant and Husain, Kabir and Sheng, Jiming and Wang, Shenshen and Murugan, Arvind},
  journal={Proceedings of the National Academy of Sciences},
  volume={117},
  number={23},
  pages={12693--12699},
  year={2020},
  publisher={National Academy of Sciences}
}

@inproceedings{meng2022rome,
  title={Locating and Editing Factual Associations in {GPT}},
  author={Meng, Kevin and Bau, David and Andonian, Alex and Belinkov, Yonatan},
  booktitle={Advances in Neural Information Processing Systems},
  year={2022}
}

@inproceedings{meng2023memit,
  title={Mass-Editing Memory in a Transformer},
  author={Meng, Kevin and Sharma, Arnab Sen and Andonian, Alex and Belinkov, Yonatan and Bau, David},
  booktitle={International Conference on Learning Representations},
  year={2023}
}

@inproceedings{
mitchell2022fast,
title={Fast Model Editing at Scale},
author={Eric Mitchell and Charles Lin and Antoine Bosselut and Chelsea Finn and Christopher D Manning},
booktitle={International Conference on Learning Representations},
year={2022}
}

@article{zweiger2025self,
  title={Self-Adapting Language Models},
  author={Zweiger, Adam and Pari, Jyothish and Guo, Han and Akyürek, Ekin and Kim, Yoon and Agrawal, Pulkit},
  journal={arXiv preprint arXiv:2506.10943},
  year={2025}
}

@article{allenzhu2023physics,
  title={Physics of Language Models: Part 3.1, Knowledge Storage and Extraction},
  author={Allen-Zhu, Zeyuan and Li, Yuanzhi},
  journal={arXiv preprint arXiv:2309.14316},
  year={2023}
}

@inproceedings{jang2022temporalwiki,
  title={{TemporalWiki}: A Lifelong Benchmark for Training and Evaluating Ever-Evolving Language Models},
  author={Jang, Joel and Ye, Seonghyeon and Lee, Changho and Yang, Sohee and Shin, Joongbo and Han, Janghoon and Kim, Gyeonghun and Seo, Minjoon},
  booktitle={Proceedings of EMNLP},
  year={2022}
}

@inproceedings{loureiro2022timelms,
  title={{TimeLMs}: Diachronic Language Models from {Twitter}},
  author={Loureiro, Daniel and Barbieri, Francesco and Neves, Leonardo and Espinosa Anke, Luis and Camacho-Collados, Jose},
  booktitle={Proceedings of ACL, System Demonstrations},
  year={2022}
}

@inproceedings{park2025temporal,
  title={Does time have its place? temporal heads: Where language models recall time-specific information},
  author={Park, Yein and Yoon, Chanwoong and Park, Jungwoo and Jeong, Minbyul and Kang, Jaewoo},
  booktitle={Proceedings of the 63rd Annual Meeting of the Association for Computational Linguistics (Volume 1: Long Papers)},
  pages={16616--16643},
  year={2025}
}

@article{dohare2024loss,
  title={Loss of Plasticity in Deep Continual Learning},
  author={Dohare, Shibhansh and Hernandez-Garcia, J. Fernando and Lan, Qingfeng and Rahman, Parash and Mahmood, A. Rupam and Sutton, Richard S.},
  journal={Nature},
  volume={632},
  pages={768--774},
  year={2024}
}

@inproceedings{hu2022lora,
  title={{LoRA}: Low-Rank Adaptation of Large Language Models},
  author={Hu, Edward J. and Shen, Yelong and Wallis, Phillip and Allen-Zhu, Zeyuan and Li, Yuanzhi and Wang, Shean and Wang, Lu and Chen, Weizhu},
  booktitle={International Conference on Learning Representations},
  year={2022}
}

@article{dhingra2022time,
  title={Time-Aware Language Models as Temporal Knowledge Bases},
  author={Dhingra, Bhuwan and Cole, Jeremy R. and Eisenschlos, Julian Martin and Gillick, Daniel and Eisenstein, Jacob and Cohen, William W.},
  journal={Transactions of the Association for Computational Linguistics},
  volume={10},
  pages={257--273},
  year={2022}
}

@inproceedings{lazaridou2021mind,
  title={Mind the Gap: Assessing Temporal Generalization in Neural Language Models},
  author={Lazaridou, Angeliki and Kuncoro, Adhiguna and Gribovskaya, Elena and Agrawal, Devang and Liska, Adam and Terzi, Tayfun and Gimenez, Mai and de Masson d'Autume, Cyprien and Kocisky, Tomas and Ruder, Sebastian and Yogatama, Dani and Cao, Kris and Young, Susannah and Blunsom, Phil},
  booktitle={Advances in Neural Information Processing Systems},
  year={2021}
}

@inproceedings{jang2022towards,
  title={Towards Continual Knowledge Learning of Language Models},
  author={Jang, Joel and Ye, Seonghyeon and Yang, Sohee and Shin, Joongbo and Han, Janghoon and Kim, Gyeonghun and Choi, Stanley Jungkyu and Seo, Minjoon},
  booktitle={International Conference on Learning Representations},
  year={2022}
}

@inproceedings{li2025tic,
  title={TiC-LM: A web-scale benchmark for time-continual LLM pretraining},
  author={Li, Jeffrey and Armandpour, Mohammadreza and Mirzadeh, Seyed Iman and Mehta, Sachin and Shankar, Vaishaal and Vemulapalli, Raviteja and Bengio, Samy and Tuzel, Oncel and Farajtabar, Mehrdad and Pouransari, Hadi and others},
  booktitle={Proceedings of the 63rd Annual Meeting of the Association for Computational Linguistics (Volume 1: Long Papers)},
  pages={32231--32273},
  year={2025}
}

@inproceedings{gururangan2020dont,
  title={Don’t stop pretraining: Adapt language models to domains and tasks},
  author={Gururangan, Suchin and Marasovi{\'c}, Ana and Swayamdipta, Swabha and Lo, Kyle and Beltagy, Iz and Downey, Doug and Smith, Noah A},
  booktitle={Proceedings of the 58th annual meeting of the association for computational linguistics},
  pages={8342--8360},
  year={2020}
}

@article{ibrahim2024simple,
  title={Simple and Scalable Strategies to Continually Pre-train Large Language Models},
  author={Ibrahim, Adam and Th{\'e}rien, Benjamin and Gupta, Kshitij and Richter, Mats L. and Anthony, Quentin and Lesort, Timoth{\'e}e and Belilovsky, Eugene and Rish, Irina},
  journal={Transactions on Machine Learning Research},
  year={2024}
}

@inproceedings{bengio2009curriculum,
  title={Curriculum learning},
  author={Bengio, Yoshua and Louradour, J{\'e}r{\^o}me and Collobert, Ronan and Weston, Jason},
  booktitle={Proceedings of the 26th annual international conference on machine learning},
  pages={41--48},
  year={2009}
}

@inproceedings{gekhman2024does,
  title={Does Fine-Tuning {LLMs} on New Knowledge Encourage Hallucinations?},
  author={Gekhman, Zorik and Yona, Gal and Aharoni, Roee and Eyal, Matan and Feder, Amir and Reichart, Roi and Herzig, Jonathan},
  booktitle={Proceedings of EMNLP},
  pages={7765--7784},
  year={2024}
}

@inproceedings{decao2021editing,
  title={Editing Factual Knowledge in Language Models},
  author={De Cao, Nicola and Aziz, Wilker and Titov, Ivan},
  booktitle={Proceedings of EMNLP},
  year={2021}
}

@inproceedings{yao2023editing,
  title={Editing Large Language Models: Problems, Methods, and Opportunities},
  author={Yao, Yunzhi and Wang, Peng and Tian, Bozhong and Cheng, Siyuan and Li, Zhoubo and Deng, Shumin and Chen, Huajun and Zhang, Ningyu},
  booktitle={Proceedings of EMNLP},
  year={2023}
}

@inproceedings{hartvigsen2023grace,
  title={Aging with {GRACE}: Lifelong Model Editing with Discrete Key-Value Adaptors},
  author={Hartvigsen, Thomas and Sankaranarayanan, Swami and Palangi, Hamid and Kim, Yoon and Ghassemi, Marzyeh},
  booktitle={Advances in Neural Information Processing Systems},
  year={2023}
}

@article{kirkpatrick2017overcoming,
  title={Overcoming catastrophic forgetting in neural networks},
  author={Kirkpatrick, James and Pascanu, Razvan and Rabinowitz, Neil and Veness, Joel and Desjardins, Guillaume and Rusu, Andrei A and Milan, Kieran and Quan, John and Ramalho, Tiago and Grabska-Barwinska, Agnieszka and others},
  journal={Proceedings of the national academy of sciences},
  volume={114},
  number={13},
  pages={3521--3526},
  year={2017},
  publisher={National Academy of Sciences}
}

@inproceedings{vu2024freshllms,
  title={{FreshLLMs}: Refreshing Large Language Models with Search Engine Augmentation},
  author={Vu, Tu and Iyyer, Mohit and Wang, Xuezhi and Constant, Noah and Wei, Jerry and Wei, Jason and Tar, Chris and Sung, Yun-Hsuan and Zhou, Denny and Le, Quoc and Luong, Thang},
  booktitle={Findings of the Association for Computational Linguistics: ACL 2024},
  pages={13697--13720},
  year={2024}
}

@inproceedings{zhao2024set,
  title={Set the Clock: Temporal Alignment of Pretrained Language Models},
  author={Zhao, Bowen and Brumbaugh, Zander and Wang, Yizhong and Hajishirzi, Hannaneh and Smith, Noah A.},
  booktitle={Findings of the Association for Computational Linguistics: ACL 2024},
  pages={15015--15040},
  year={2024}
}

@inproceedings{gurnee2024language,
  title={Language Models Represent Space and Time},
  author={Gurnee, Wes and Tegmark, Max},
  booktitle={International Conference on Learning Representations},
  year={2024}
}

@article{pilchen2026understanding,
  title={Understanding Data Temporality Impact on Large Language Models Pre-training},
  author={Pilchen, Hippolyte and Fabre, Romain and Signe Talla, Franck and Perez, Patrick and Grave, Edouard},
  journal={arXiv preprint arXiv:2605.22769},
  year={2026}
}

@inproceedings{wu2021when,
  title={When Do Curricula Work?},
  author={Wu, Xiaoxia and Dyer, Ethan and Neyshabur, Behnam},
  booktitle={International Conference on Learning Representations},
  year={2021}
}

@inproceedings{jin2022lifelong,
  title={Lifelong Pretraining: Continually Adapting Language Models to Emerging Corpora},
  author={Jin, Xisen and Zhang, Dejiao and Zhu, Henghui and Xiao, Wei and Li, Shang-Wen and Wei, Xiaokai and Arnold, Andrew and Ren, Xiang},
  booktitle={Proceedings of NAACL},
  pages={4764--4780},
  year={2022}
}

@inproceedings{ke2023continual,
  title={Continual Pre-training of Language Models},
  author={Ke, Zixuan and Shao, Yijia and Lin, Haowei and Konishi, Tatsuya and Kim, Gyuhak and Liu, Bing},
  booktitle={International Conference on Learning Representations},
  year={2023}
}

@inproceedings{toneva2019empirical,
  title={An Empirical Study of Example Forgetting during Deep Neural Network Learning},
  author={Toneva, Mariya and Sordoni, Alessandro and Tachet des Combes, Remi and Trischler, Adam and Bengio, Yoshua and Gordon, Geoffrey J.},
  booktitle={International Conference on Learning Representations},
  year={2019}
}

@inproceedings{tirumala2022memorization,
  title={Memorization Without Overfitting: Analyzing the Training Dynamics of Large Language Models},
  author={Tirumala, Kushal and Markosyan, Aram H. and Zettlemoyer, Luke and Aghajanyan, Armen},
  booktitle={Advances in Neural Information Processing Systems},
  year={2022}
}

@inproceedings{mitchell2022memory,
  title={Memory-Based Model Editing at Scale},
  author={Mitchell, Eric and Lin, Charles and Bosselut, Antoine and Manning, Christopher D. and Finn, Chelsea},
  booktitle={International Conference on Machine Learning},
  pages={15817--15831},
  year={2022}
}

@inproceedings{fang2025alphaedit,
  title={{AlphaEdit}: Null-Space Constrained Knowledge Editing for Language Models},
  author={Fang, Junfeng and Jiang, Houcheng and Wang, Kun and Ma, Yunshan and Shi, Jie and Wang, Xiang and He, Xiangnan and Chua, Tat-Seng},
  booktitle={International Conference on Learning Representations},
  year={2025}
}

@inproceedings{wang2024wise,
  title={{WISE}: Rethinking the Knowledge Memory for Lifelong Model Editing of Large Language Models},
  author={Wang, Peng and Li, Zexi and Zhang, Ningyu and Xu, Ziwen and Yao, Yunzhi and Jiang, Yong and Xie, Pengjun and Huang, Fei and Chen, Huajun},
  booktitle={Advances in Neural Information Processing Systems},
  year={2024}
}

@inproceedings{ovadia2024fine,
  title={Fine-Tuning or Retrieval? Comparing Knowledge Injection in {LLMs}},
  author={Ovadia, Oded and Brief, Menachem and Mishaeli, Moshik and Elisha, Oren},
  booktitle={Proceedings of EMNLP},
  pages={237--250},
  year={2024}
}

@inproceedings{kim2025knowledge,
  title={Knowledge Entropy Decay during Language Model Pretraining Hinders New Knowledge Acquisition},
  author={Kim, Jiyeon and Lee, Hyunji and Cho, Hyowon and Jang, Joel and Hwang, Hyeonbin and Won, Seungpil and Ahn, Youbin and Lee, Dohaeng and Seo, Minjoon},
  booktitle={International Conference on Learning Representations},
  year={2025}
}

@inproceedings{yang2025synthetic,
  title={Synthetic Continued Pretraining},
  author={Yang, Zitong and Band, Neil and Li, Shuangping and Cand{\`e}s, Emmanuel and Hashimoto, Tatsunori},
  booktitle={International Conference on Learning Representations},
  year={2025}
}

@inproceedings{liu2025drop,
  title={Drop Dropout on Single-Epoch Language Model Pretraining},
  author={Liu, Houjun and Bauer, John and Manning, Christopher D.},
  booktitle={Findings of the Association for Computational Linguistics: ACL 2025},
  year={2025}
}

@inproceedings{scialom2022fine,
  title={Fine-tuned Language Models are Continual Learners},
  author={Scialom, Thomas and Chakrabarty, Tuhin and Muresan, Smaranda},
  booktitle={Proceedings of EMNLP},
  pages={6107--6122},
  year={2022}
}

@article{luo2023empirical,
  title={An Empirical Study of Catastrophic Forgetting in Large Language Models During Continual Fine-tuning},
  author={Luo, Yun and Yang, Zhen and Meng, Fandong and Li, Yafu and Zhou, Jie and Zhang, Yue},
  journal={arXiv preprint arXiv:2308.08747},
  year={2023}
}

@inproceedings{gupta2025encore,
  title={Lifelong Knowledge Editing requires Better Regularization},
  author={Gupta, Akshat and Prateepamornkul, Phudish and Lu, Maochuan and Alaa, Ahmed and Hartvigsen, Thomas and Anumanchipalli, Gopala},
  booktitle={Findings of the Association for Computational Linguistics: EMNLP 2025},
  year={2025}
}

@article{team2024gemma,
  title={Gemma 2: Improving open language models at a practical size},
  author={Team, Gemma},
  journal={arXiv preprint arXiv:2408.00118},
  year={2024}
}

@article{su2024roformer,
  title={Roformer: Enhanced transformer with rotary position embedding},
  author={Su, Jianlin and Ahmed, Murtadha and Lu, Yu and Pan, Shengfeng and Bo, Wen and Liu, Yunfeng},
  journal={Neurocomputing},
  volume={568},
  pages={127063},
  year={2024},
  publisher={Elsevier}
}

@inproceedings{kingma2015adam,
  title={Adam: A Method for Stochastic Optimization},
  author={Kingma, Diederik P. and Ba, Jimmy},
  booktitle={International Conference on Learning Representations},
  year={2015}
}

@article{chen2023symbolic,
  title={Symbolic discovery of optimization algorithms},
  author={Chen, Xiangning and Liang, Chen and Huang, Da and Real, Esteban and Wang, Kaiyuan and Pham, Hieu and Dong, Xuanyi and Luong, Thang and Hsieh, Cho-Jui and Lu, Yifeng and others},
  journal={Advances in neural information processing systems},
  volume={36},
  pages={49205--49233},
  year={2023}
}

@article{clark2018think,
  title={Think you have solved question answering? try arc, the ai2 reasoning challenge},
  author={Clark, Peter and Cowhey, Isaac and Etzioni, Oren and Khot, Tushar and Sabharwal, Ashish and Schoenick, Carissa and Tafjord, Oyvind},
  journal={arXiv preprint arXiv:1803.05457},
  year={2018}
}

@inproceedings{bisk2020piqa,
  title={Piqa: Reasoning about physical commonsense in natural language},
  author={Bisk, Yonatan and Zellers, Rowan and Le Bras, Ronan and Gao, Jianfeng and Choi, Yejin},
  booktitle={Proceedings of the AAAI Conference on Artificial Intelligence},
  volume={34},
  number={05},
  pages={7432--7439},
  year={2020}
}

@article{johnson2025exploring,
  title={Exploring collaborative GenAI agents in synchronous group settings: eliciting team perceptions and design considerations for the future of work},
  author={Johnson, Janet G and Peralta, Macarena and Kaur, Mansanjam and Huang, Ruijie Sophia and Zhao, Sheng and Guan, Ruijia and Rajaram, Shwetha and Nebeling, Michael},
  journal={Proceedings of the ACM on Human-Computer Interaction},
  volume={9},
  number={7},
  pages={1--33},
  year={2025},
  publisher={ACM New York, NY, USA}
}

@inproceedings{zellers2019hellaswag,
  title={Hellaswag: Can a machine really finish your sentence?},
  author={Zellers, Rowan and Holtzman, Ari and Bisk, Yonatan and Farhadi, Ali and Choi, Yejin},
  booktitle={Proceedings of the 57th annual meeting of the association for computational linguistics},
  pages={4791--4800},
  year={2019}
}

@article{penedo2024fineweb,
  title={The fineweb datasets: Decanting the web for the finest text data at scale},
  author={Penedo, Guilherme and Kydl{\'\i}{\v{c}}ek, Hynek and Lozhkov, Anton and Mitchell, Margaret and Raffel, Colin and Von Werra, Leandro and Wolf, Thomas and others},
  journal={Advances in Neural Information Processing Systems},
  volume={37},
  pages={30811--30849},
  year={2024}
}
\bibliographystyle{iclr2027_conference}

\clearpage
\appendix
\section*{Appendix}
\section{Data, Training and Query Protocol}
\label{app:protocol}

This appendix gives the detailed settings behind \Sect{sec:setup} and \Sect{sec:shape}: how the biographies are generated, how the models are trained, how a query is scored, and how an update is run and measured.

\paragraph{Biographies.}
Names and attribute values come from the vocabulary lists of \citet{allenzhu2023physics}.
Each person appears 1{,}000 times between 2000 and 2024, and each appearance is one record that states the six attributes.
Birthday and birth city never change.
University and field change 3 times over a lifetime (median; mean 2.8), employer and employer city 10 times (mean 9.9), at random points of the 1{,}000 appearances.

\begin{wrapfigure}{r}{0.45\linewidth}
\vspace{-10pt}
\centering
\includegraphics[width=\linewidth]{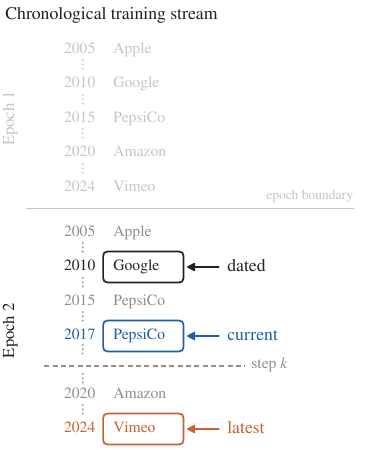}
\caption{Targets of the three kinds of query in \Fig{fig:tracking-full}: a value in the queried year, the value in the batch trained at step $k$, and the final value. 
}
\label{fig:probes}
\vspace{-10pt}
\end{wrapfigure}

In a chronological stream, consecutive records of the same person usually contain the same value.
In CD, the values are unchanged in about 99.0\% of consecutive records.
After shuffling, consecutive records can come from very different years, so these numbers drop to around 20.0\%.
So in a chronological stream the year says almost nothing about a person's current value; 
\Sect{sec:time} relies on this, and \Apx{app:temporal-risk} turns it into a bound.
1{,}000 individuals in the corpus are held out from training and used only for injection updates; 
A timestamped record starts with ``In \{year\},''; the half-timestamped corpora drop this prefix from a random half of the records, the same half in CD\half{} and SD\half{}, so that the two differ only in order (\Tbl{tab:datasets}).

\paragraph{Pre-training.}
The three scales share the GPT-2 architecture with rotary position embeddings \citep{su2024roformer} and the GPT-2 tokenizer (\Tbl{tab:arch}).
Every model trains on batches of 2{,}560 records, with a 5\% linear warmup, a cosine decay from the peak learning rate of \Tbl{tab:arch} to a tenth of it, weight decay 0.1 and gradient clipping at 1.0.
For the main results in \Sect{sec:shape}, we train the models for five epochs, by which point query accuracy has converged already.

\begin{figure}[t]
\centering
\includegraphics[width=\linewidth]{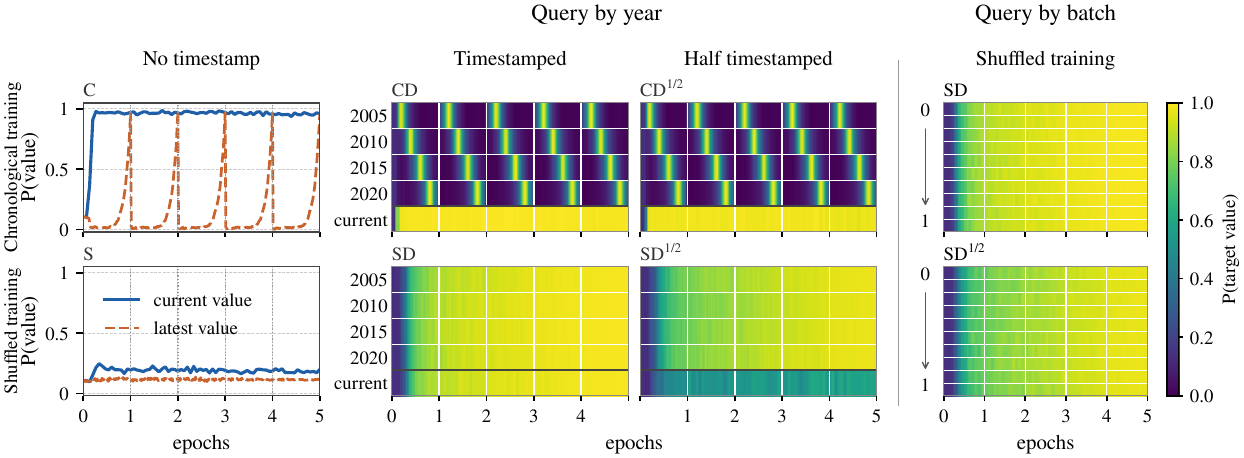}
\caption{
Query accuracy of the six 110M models over five epochs of training, the full version of \Fig{fig:tracking}.
Rows of panels compare chronological and shuffled training, columns compare timestamp settings, and the right column queries SD and SD$^{1/2}$ by batch as in \Fig{fig:tracking}c.
Rows 2005--2020 ask with a year (``In 2005, \ldots'') and count any value the person held in that year; the \emph{current} row asks about records of the batch the model has just trained on, in the form each was trained in.
In the two panels on the left, the solid curve is this current score for the undated models and the dashed curve the probability of the final value, held in 2024.
}
\label{fig:tracking-full}
\end{figure}

\paragraph{Queries.}
A query prompts the model with a person's name and an attribute, cut just before the value, and reads the probability of every value in the attribute's vocabulary.
A dated query starts with ``In \{year\},'' and is correct if the answer is any value the person held during that year.
In contrast, an undated query gives no year, and is correct if the answer is the latest value (See \Fig{fig:probes} about ``latest'').
The year rows (2005--2020) queries 200 people per time-varying attribute about a fixed year, every 125 steps.
The queries keep unchanged for different settings.
The current row asks, at every checkpoint (every 625 steps), about 200 records of the batch the model has just trained on, in the form each record was trained in.
The query-by-batch rows of \Fig{fig:tracking}c ask in the same way about 200 records from each of ten batches spaced evenly over the SD stream.
\Fig{fig:tracking-full} extends \Fig{fig:tracking} to all six corpora.

\begin{table}[t]
\centering
\footnotesize
\begin{minipage}[b]{0.34\linewidth}
\centering
\caption{Pre-training datasets. C/S: chronological/shuffled; D: timestamped; \half: half timestamped.
}
\label{tab:datasets}
\scriptsize
\setlength{\tabcolsep}{4pt}
\begin{tabular}{lcc}
\toprule
Timestamps & Chronological & Shuffled \\
\midrule
0\%   & C       & S       \\
50\%  & CD\half & SD\half \\
100\% & CD      & SD      \\
\bottomrule
\end{tabular}
\end{minipage}\hfill
\begin{minipage}[b]{0.62\linewidth}
\centering
\captionsetup{width=0.9\linewidth}
\caption{Model scales, all with a context of 512 tokens. Records: per epoch; LR: peak learning rate. Other settings in \Apx{app:protocol}.
}
\label{tab:arch}
\scriptsize
\setlength{\tabcolsep}{3pt}
\begin{tabular}{lrrrrrr}
\toprule
Scale & Layers & Heads & Hidden & Records & Tokens & LR ($10^{-4}$) \\
\midrule
110M & 12 & 12 & 768  & 24M  & 10.88B & 6   \\
350M & 24 & 16 & 1024 & 68M  & 30.80B & 3   \\
682M & 32 & 20 & 1280 & 134M & 60.68B & 2.5 \\
\bottomrule
\end{tabular}
\end{minipage}
\end{table}

\paragraph{Updates.}
An update trains a rank-16 LoRA adapter \citep{hu2022lora}, with $\alpha=16$, on every attention and MLP projection of every layer (84 adapters at 110M), using Adam \citep{kingma2015adam} without weight decay at a fixed learning rate to the pre-training peak (1/60, e.g. $10^{-5}$ at 110M), no weight decay.
The training sentence states the new value in one of the attribute's sentence templates; for timestamped models it starts with ``In 2024,'', and for half-timestamped models a random half of the updates do.
An \emph{overwrite} update targets a person the model was trained on and an \emph{injection} update targets the held-out person during training.
Either succeeds when the new value's probability passes that of its competitor (i.e. the strongest value the model held before updating).

The learning gap $\Dist$ is the log-probability gap to that competitor before the update.
Each model receives 480 updates, 40 \emph{overwrite} updates and 40 \emph{injection} updates per attribute.
For each attribute we sample the individuals so that different people are equally represented.
\Tbl{tab:steps} reports medians over all overwrite updates or all injection updates of a model.
Forgetting is measured on facts of other people, as the percentage whose answer the update changes.

\paragraph{Updating criteria.}
\label{app:strict}
Our results do not depend on the stopping criterion used for an update.
On 120 updates from each 110M pair in \Tbl{tab:steps}, stopping when the new value passes the model's strongest previous value gives the same conclusion as requiring it to become the top-1 value over the full attribute vocabulary (\Tbl{tab:strict}).
The same conclusion also holds with two stricter thresholds: a 3-nat log-probability margin over the competitor or $P_{\mathrm{new}}\geq 0.5$.

\begin{table}[h]
\centering
\footnotesize
\caption{Update cost of the 110M pairs under three stricter success criteria, which are the new value is the top-1 over the attribute's whole vocabulary, its log-probability exceeds the competitor's by 3 and its probability $P_{\mathrm{new}}$ reaches 0.5. Cells represent median steps, chronological / shuffled, and the share of steps the chronological model saves (bold: paired Wilcoxon $p<0.05$). 60 \emph{overwrites} and 60 \emph{injections} updates per pair.}
\label{tab:strict}
\scriptsize
\setlength{\tabcolsep}{3pt}
\resizebox{\linewidth}{!}{\begin{tabular}{llcccccccccccc}
\toprule
 & & \multicolumn{6}{c}{Overwrite} & \multicolumn{6}{c}{Inject} \\
\cmidrule(lr){3-8}\cmidrule(lr){9-14}
 & & \multicolumn{2}{c}{Vocabulary top-1} & \multicolumn{2}{c}{Margin $\ge 3$} & \multicolumn{2}{c}{$P_{\mathrm{new}}\ge 0.5$} & \multicolumn{2}{c}{Vocabulary top-1} & \multicolumn{2}{c}{Margin $\ge 3$} & \multicolumn{2}{c}{$P_{\mathrm{new}}\ge 0.5$} \\
\cmidrule(lr){3-4}\cmidrule(lr){5-6}\cmidrule(lr){7-8}\cmidrule(lr){9-10}\cmidrule(lr){11-12}\cmidrule(lr){13-14}
Scale & Pair & steps & saved & steps & saved & steps & saved & steps & saved & steps & saved & steps & saved \\
\midrule
\multirow{3}{*}{110M} & C / S & 53.0 / 89.0 & \textbf{$+$40\%} & 63.5 / 108 & \textbf{$+$41\%} & 59.5 / 99.5 & \textbf{$+$40\%} & 38.0 / 57.0 & \textbf{$+$33\%} & 46.5 / 78.5 & \textbf{$+$41\%} & 47.0 / 71.5 & \textbf{$+$34\%} \\
 & CD / SD & 51.5 / 75.0 & \textbf{$+$31\%} & 63.0 / 88.5 & \textbf{$+$29\%} & 57.5 / 80.5 & \textbf{$+$29\%} & 40.0 / 52.0 & \textbf{$+$23\%} & 53.0 / 66.5 & \textbf{$+$20\%} & 50.0 / 62.5 & \textbf{$+$20\%} \\
 & CD\half{} / SD\half{} & 54.5 / 88.0 & \textbf{$+$38\%} & 67.0 / 103 & \textbf{$+$35\%} & 61.0 / 96.5 & \textbf{$+$37\%} & 31.5 / 61.5 & \textbf{$+$49\%} & 40.5 / 84.5 & \textbf{$+$52\%} & 39.5 / 80.0 & \textbf{$+$51\%} \\
\bottomrule
\end{tabular}
}
\end{table}

\section{Generalists and Specialists: Derivations and Measurements}
\label{app:formation}

\subsection{The Predictive Value of Timestamps}
\label{app:temporal-risk}

This appendix derives \Equ{eq:temporal-objective} and \Equ{eq:temporal-risk} in \Sect{sec:time}.
It shows how much the year helps prediction within a training window.
Chronological and shuffled training have the same global objective, but the year is much less useful in a chronological window than in a shuffled one.

\begin{figure}[h]
\centering
\includegraphics[width=0.6\linewidth]{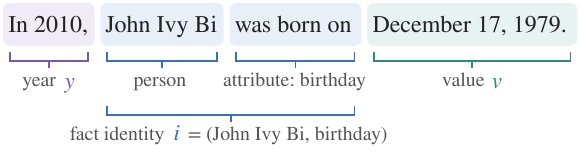}
\caption{Mapping the example in \Sect{sec:setup} to the fact-level variables. The person and attribute jointly identify the fact $i$; $y=2010$ is the record's timestamp, and $v=\text{December 17, 1979}$ is its value. Uppercase $I,Y,V$ denote the variables and lowercase $i,y,v$ their observed values. Here the attribute is time-invariant; for a changing attribute such as employer, the same $i$ can have different values $v$ in different years $y$.}
\label{fig:temporal-example}
\end{figure}

\paragraph{Step 1: the global objective is the same for both orders.}
Pretraining minimizes the negative log-likelihood of whole records, whereas a query tests the value of one fact.
For a tokenized record $x=(x_1,\ldots,x_T)$ whose value tokens sit at positions $\mathcal V(x)$, the record loss splits exactly into a value part and a remainder,
\begin{equation}
\ell_{\mathrm{pre}}(\theta;x)
=-\sum_{j=1}^{T}\log p_\theta(x_j\mid x_{<j})
=\underbrace{-\sum_{j\in\mathcal V(x)}\log p_\theta(x_j\mid x_{<j})}_{\ell_{\mathrm{value}}(\theta;x)}
+\underbrace{-\sum_{j\notin\mathcal V(x)}\log p_\theta(x_j\mid x_{<j})}_{\ell_{\mathrm{other}}(\theta;x)},
\label{eq:temporal-record-loss}
\end{equation}
and for a value span $v$ after context $c$ the value part equals $-\log p_\theta(v\mid c)$ by the chain rule.
We study the value part, since it trains the associations a query measures, and represent its context by the fact identity $I$ and the year $Y$ (\Fig{fig:temporal-example}); if further context $C$ is kept, every quantity below is also conditioned on $C$ and nothing else changes.
Let $\mathcal D=\{(i_m,y_m,v_m)\}_{m=1}^{N}$ be the fact-level records.
Their empirical distribution and value-prediction loss are
\begin{equation}
P(i,y,v)=\frac1N\sum_{m=1}^{N}\mathbf1[(i_m,y_m,v_m)=(i,y,v)],\qquad
\bar L(\theta)=\mathbb E_P[-\log p_\theta(V\mid I,Y)].
\label{eq:temporal-global}
\end{equation}
Permuting the records changes neither $P$ nor $\bar L(\theta)$ at any $\theta$, so chronological and shuffled training don't differ in their global objective.

\paragraph{Step 2: the local objective depends on the order.}
For an ordering $Z_1,\ldots,Z_N$ with $Z_n=(I_n,Y_n,V_n)$, the empirical distribution of the $W$ consecutive records from position $t$ is
\begin{equation}
Q_{t,W}(i,y,v)=\frac1W\sum_{n=t}^{t+W-1}\mathbf1[Z_n=(i,y,v)],
\qquad 1\le t\le N-W+1.
\label{eq:temporal-window}
\end{equation}
The horizon $W$ is an analysis scale, not necessarily a batch size.
Two orderings share $Q_{1,N}=P$ but can have different $Q_{t,W}$ for $W<N$.
We fix a window and write $Q=Q_{t,W}$, with every conditional quantity evaluated on its support.

\paragraph{Step 3: the best predictor with and without the year.}
A year-independent predictor $r(V\mid I)$ has loss $-\sum_{i,v}Q(i,v)\log r(v\mid i)$.
Writing $-\log r(v\mid i)=-\log Q(v\mid i)+\log[Q(v\mid i)/r(v\mid i)]$ and using $Q(i,v)=Q(i)Q(v\mid i)$,
\begin{align}
L_Q^{\mathrm{blind}}(r)
&=-\sum_{i,y,v}Q(i,y,v)\log r(v\mid i)
=-\sum_{i,v}Q(i,v)\log r(v\mid i)\nonumber\\
&=-\sum_i Q(i)\sum_v Q(v\mid i)\log Q(v\mid i)\nonumber\\
&\quad+\sum_i Q(i)\sum_v Q(v\mid i)\log\frac{Q(v\mid i)}{r(v\mid i)}\nonumber\\
&=H_Q(V\mid I)+\mathbb E_{I\sim Q}
D_{\mathrm{KL}}\!\left(Q(V\mid I)\,\Vert\,r(V\mid I)\right).
\label{eq:temporal-blind-kl}
\end{align}
Here $H$ is the entropy in bits, $H_Q(V\mid I)=\sum_i Q(i)H(Q(V\mid i))$ is the uncertainty in the value once the fact is known, and the second term is a Kullback--Leibler (KL) divergence, nonnegative and zero at $r^*(v\mid i)=Q(v\mid i)$, the window's mixture of the fact's values across years.
The same steps for a year-conditioned predictor $p(V\mid I,Y)$, with $Q(i,y,v)=Q(i,y)Q(v\mid i,y)$, give
\begin{align}
L_Q^{\mathrm{dated}}(p)
&=-\sum_{i,y}Q(i,y)\sum_v Q(v\mid i,y)\log p(v\mid i,y)\nonumber\\
&=-\sum_{i,y}Q(i,y)\sum_v Q(v\mid i,y)\log Q(v\mid i,y)\nonumber\\
&\quad+\sum_{i,y}Q(i,y)\sum_v Q(v\mid i,y)\log\frac{Q(v\mid i,y)}{p(v\mid i,y)}\nonumber\\
&=H_Q(V\mid I,Y)+\mathbb E_{I,Y\sim Q}
D_{\mathrm{KL}}\!\left(Q(V\mid I,Y)\,\Vert\,p(V\mid I,Y)\right),
\label{eq:temporal-dated-kl}
\end{align}
where $H_Q(V\mid I,Y)=\sum_{i,y}Q(i,y)H(Q(V\mid i,y))$ is the uncertainty once fact and year are known, and the divergence is zero at $p^*(v\mid i,y)=Q(v\mid i,y)$.
Over unrestricted predictors, therefore,
\begin{equation}
\inf_r L_Q^{\mathrm{blind}}(r)=H_Q(V\mid I),\qquad
\inf_p L_Q^{\mathrm{dated}}(p)=H_Q(V\mid I,Y).
\label{eq:temporal-bayes-optima}
\end{equation}
These are the minimum possible losses for the two predictors within the window.
The dated loss is zero only when the fact and year uniquely determine the value.

\paragraph{Step 4: the gap is a conditional mutual information.}
Subtracting the two optima,
\begin{align}
R_{t,W}
&:=\inf_r L_Q^{\mathrm{blind}}(r)-\inf_p L_Q^{\mathrm{dated}}(p)\nonumber\\
&=H_Q(V\mid I)-H_Q(V\mid I,Y)\nonumber\\
&=\sum_{i,y,v}Q(i,y,v)\log\frac{Q(v\mid i,y)}{Q(v\mid i)}
=\operatorname{MI}_Q(V;Y\mid I),
\label{eq:temporal-gap-proof}
\end{align}
where the last step uses $Q(v,y\mid i)=Q(y\mid i)Q(v\mid i,y)$.
The identity holds for every ordering and every window: $R_{t,W}$ is the minimum loss a predictor pays for ignoring the year, and it is positive exactly when the value distribution of a fact depends on the year.

\paragraph{Step 5: an upper bound on the gap.}
\label{para: near_zero}
In a chronological window a fact's value is almost constant even though its year changes.
For each fact $i$ with $Q(i)>0$, let $v_i^\star\in\arg\max_v Q(v\mid i)$ be a most probable value, $\varepsilon_i=1-Q(v_i^\star\mid i)$ the probability of any other value, and $K_i=|\operatorname{supp}Q(V\mid i)|$ the number of values.
For $\varepsilon_i>0$, normalize the remaining mass as $\widetilde Q_i(v)=Q(v\mid i)/\varepsilon_i$ for $v\ne v_i^\star$.
Separating the modal value from this tail gives
\begin{align}
H_Q(V\mid I=i)
&=-(1-\varepsilon_i)\log(1-\varepsilon_i)
-\sum_{v\ne v_i^\star}\varepsilon_i\widetilde Q_i(v)
\log\!\left(\varepsilon_i\widetilde Q_i(v)\right)\nonumber\\
&=h_2(\varepsilon_i)+\varepsilon_i H(\widetilde Q_i)
\le h_2(\varepsilon_i)+\varepsilon_i\log(K_i-1),
\label{eq:temporal-persistence-entropy}
\end{align}
where $h_2(\varepsilon)=-(1-\varepsilon)\log(1-\varepsilon)-\varepsilon\log\varepsilon$ is the binary entropy and the inequality uses the tail's support of at most $K_i-1$ values; a fact with $K_i=1$ has $\varepsilon_i=0$ and contributes zero.
Since $H_Q(V\mid I,Y)\ge0$, averaging over facts gives
\begin{equation}
0\le R_{t,W}\le H_Q(V\mid I)
\le B_{t,W}:=\sum_{i:K_i\ge2}Q(i)
\left[h_2(\varepsilon_i)+\varepsilon_i\log(K_i-1)\right].
\label{eq:temporal-persistence-bound}
\end{equation}
The bound holds for either order.
Under chronological training, values are highly persistent within a window, so $\varepsilon_i$ is close to zero and the bound becomes small.

Empirically, for the 110M CD model, $R_{t,W}\leq B_{t,W}=0.01$ bits per record, while the corresponding shuffled window has $R_{t,W}=0.69$ bits, at least $69\times$ larger.
Thus, the year provides almost no predictive value in a chronological window but much more in a shuffled one.

\subsection{Tracing Timestamp Use with Activation Patching}
\label{app:time-patching}

\Sect{sec:time} states that a specialist reads the year but does not rely on it.
To test how each model uses the year during inference, we use 115 dated questions whose historical answer differs from the 2024 answer.
For each question, we run the model twice: once with the historical year and once with 2024.
We then patch one activation from the 2024 run into the historical run and ask whether this makes the model answer more like it was queried about 2024.

\begin{figure}[t]
\centering
\includegraphics[width=\linewidth]{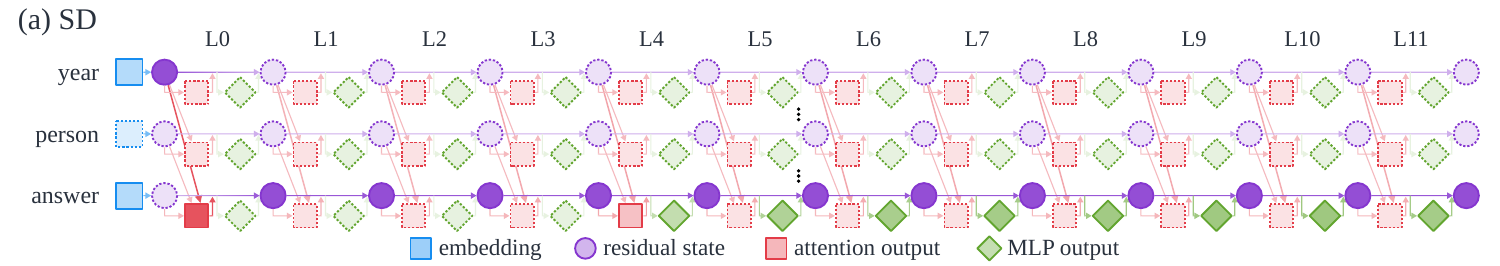}\\[2pt]
\includegraphics[width=\linewidth]{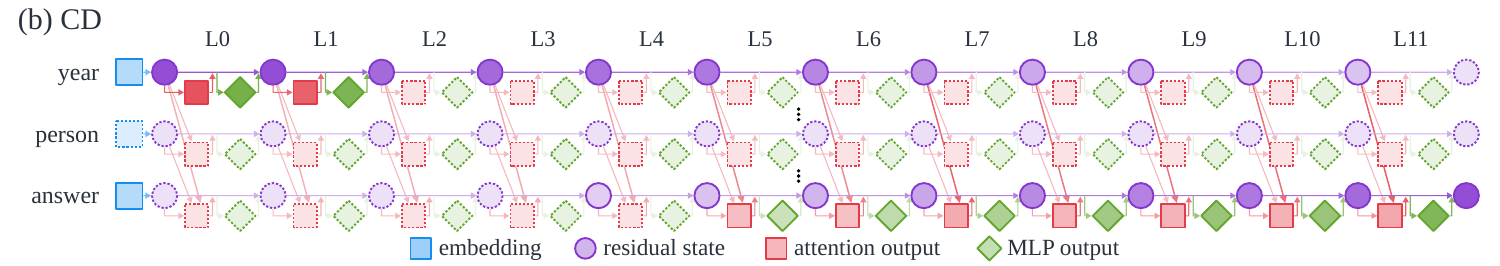}
\caption{
Activation patching of the timestamp in the 110M SD (a) and CD (b).
Rows are three positions of a dated question (the year, the person's name and the answer position) and columns the twelve layers; each symbol is one activation: embedding, residual state, attention output or MLP output.
A darker symbol recovers more of the model's own year effect when that activation is copied from the same question asked about 2024; colors compare locations within one model.
SD recovers the year at the answer position in the first attention block, whereas in CD the year effect builds up gradually through depth.
}
\label{fig:time-patching}
\end{figure}

We patch activations at the year token, the person's name, or the answer position, across different layers and modules.
The patching score measures how much of the original shift from the historical answer to the 2024 answer is recovered by a single patch.
It therefore traces how year information flows through the model during inference.
\Fig{fig:time-patching} reports the median over the questions on which the year moves the answer by more than 1.4 bits.

Both models process the year during inference, but they use it very differently.
SD uses the year strongly and early, while CD picks it up only gradually and gives it much less weight.
If CD relied on the year to predict the value, it would behave like a generalist and answer dated queries across years accurately.
Instead, chronological training places the model in a parameter region where the year is processed but is not needed much for value prediction.

\section{Update Cost: Full Results and the Step Law}
\label{app:faster}

\subsection{Update Cost by Rate of Change and Across Seeds}
\label{app:tiers}

\Tbl{tab:steps_full} extends \Tbl{tab:steps} by splitting time-varying facts into slow- and fast-changing groups and by adding the Mix-C/Mix-S pair from \Sect{sec:frontier}.
\Tbl{tab:seeds} repeats the 110M C/S comparison with three additional random initializations.
Across the four runs, chronological training saves 44--52\% of the update steps, showing that the result is robust to initialization.

\begin{table}[h]
\centering
\footnotesize
\caption{Update cost by how often a fact changes: the time-varying facts of \Tbl{tab:steps} split into slow and fast facts, with the Mix pair (\Sect{sec:frontier}) added.
Each ``steps'' cell gives the median steps of the chronological and shuffled models.
The ``saved'' column is the percentage of update steps that a chronological model saves over the corresponding shuffled model; bold values have a paired Wilcoxon $p<0.05$.}
\label{tab:steps_full}
\scriptsize
\setlength{\tabcolsep}{4.5pt}
\renewcommand{\arraystretch}{0.95}
\begin{tabular}{cllcccc}
\toprule
 & & & \multicolumn{2}{c}{Slow} & \multicolumn{2}{c}{Fast} \\
\cmidrule(lr){4-5}\cmidrule(lr){6-7}
 & Scale & Pair & steps & saved & steps & saved \\
\midrule
\multirow{10}{*}{\rotatebox{90}{Overwrite}} & \multirow{4}{*}{110M} & C / S & 49.0 / 91.5 & \textbf{$+$46\%} & 39.0 / 86.0 & \textbf{$+$55\%} \\
 &  & CD / SD & 52.5 / 69.0 & \textbf{$+$24\%} & 38.0 / 71.0 & \textbf{$+$46\%} \\
 &  & CD\half{} / SD\half{} & 51.0 / 74.5 & \textbf{$+$32\%} & 36.0 / 72.5 & \textbf{$+$50\%} \\
 &  & Mix-C / Mix-S & 46.5 / 70.5 & \textbf{$+$34\%} & 45.0 / 64.5 & \textbf{$+$30\%} \\
\cmidrule(lr){2-7}
 & \multirow{3}{*}{350M} & C / S & 47.0 / 118 & \textbf{$+$60\%} & 36.0 / 118 & \textbf{$+$70\%} \\
 &  & CD / SD & 54.0 / 81.0 & \textbf{$+$33\%} & 41.0 / 86.5 & \textbf{$+$53\%} \\
 &  & CD\half{} / SD\half{} & 49.5 / 94.5 & \textbf{$+$48\%} & 35.0 / 92.5 & \textbf{$+$62\%} \\
\cmidrule(lr){2-7}
 & \multirow{3}{*}{682M} & C / S & 45.5 / 114 & \textbf{$+$60\%} & 36.5 / 121 & \textbf{$+$70\%} \\
 &  & CD / SD & 50.0 / 80.0 & \textbf{$+$38\%} & 44.0 / 84.0 & \textbf{$+$48\%} \\
 &  & CD\half{} / SD\half{} & 48.0 / 98.0 & \textbf{$+$51\%} & 37.5 / 103 & \textbf{$+$64\%} \\
\midrule
\multirow{10}{*}{\rotatebox{90}{Inject}} & \multirow{4}{*}{110M} & C / S & 37.0 / 62.5 & \textbf{$+$41\%} & 24.5 / 54.0 & \textbf{$+$55\%} \\
 &  & CD / SD & 38.0 / 49.0 & \textbf{$+$22\%} & 23.0 / 46.5 & \textbf{$+$51\%} \\
 &  & CD\half{} / SD\half{} & 34.0 / 55.0 & \textbf{$+$38\%} & 21.0 / 51.0 & \textbf{$+$59\%} \\
 &  & Mix-C / Mix-S & 36.0 / 50.0 & \textbf{$+$28\%} & 33.0 / 44.0 & \textbf{$+$25\%} \\
\cmidrule(lr){2-7}
 & \multirow{3}{*}{350M} & C / S & 32.0 / 67.5 & \textbf{$+$53\%} & 19.0 / 59.0 & \textbf{$+$68\%} \\
 &  & CD / SD & 31.5 / 50.0 & \textbf{$+$37\%} & 20.0 / 53.5 & \textbf{$+$63\%} \\
 &  & CD\half{} / SD\half{} & 29.0 / 57.0 & \textbf{$+$49\%} & 17.0 / 55.5 & \textbf{$+$69\%} \\
\cmidrule(lr){2-7}
 & \multirow{3}{*}{682M} & C / S & 27.0 / 71.0 & \textbf{$+$62\%} & 18.0 / 60.0 & \textbf{$+$70\%} \\
 &  & CD / SD & 29.5 / 52.0 & \textbf{$+$43\%} & 21.0 / 50.5 & \textbf{$+$58\%} \\
 &  & CD\half{} / SD\half{} & 28.0 / 62.0 & \textbf{$+$55\%} & 18.0 / 55.0 & \textbf{$+$67\%} \\
\bottomrule
\end{tabular}
\renewcommand{\arraystretch}{1.0}
\end{table}

\begin{table}[h]
\centering
\small
\caption{Temporal order saves about half of the steps on time-varying facts in each of four independently trained 110M C / S pairs, each with its own population and initialization. Cells read as in \Tbl{tab:steps}: median steps on time-varying facts, chronological / shuffled, and the share of steps temporal order saves.}
\label{tab:seeds}
\setlength{\tabcolsep}{5pt}
\begin{tabular}{lcccc}
\toprule
 & \multicolumn{2}{c}{Overwrite} & \multicolumn{2}{c}{Inject} \\
\cmidrule(lr){2-3}\cmidrule(lr){4-5}
Seed & steps & saved & steps & saved \\
\midrule
0 (main) & 44 / 89 & \textbf{$+$51\%} & 29 / 57.5 & \textbf{$+$50\%} \\
1 & 46 / 89.5 & \textbf{$+$49\%} & 29.5 / 61 & \textbf{$+$52\%} \\
2 & 48 / 89 & \textbf{$+$46\%} & 32 / 57 & \textbf{$+$44\%} \\
3 & 44 / 89 & \textbf{$+$51\%} & 28 / 55.5 & \textbf{$+$50\%} \\
\bottomrule
\end{tabular}
\end{table}

\subsection{Other Optimizers, Learning Rates and Full Fine-Tuning}
\label{app:robust}

The update advantage of specialists on time-varying facts is robust to how the update is performed.
\Tbl{tab:robust} shows the same pattern across optimizers, learning rates, LoRA ranks, and full fine-tuning: chronological models consistently require fewer update steps than shuffled models.
The saving is about $50\%$ under Adam \citep{kingma2015adam} and Lion \citep{chen2023symbolic} , and is even larger under SGD.
The step law in \Sect{sec:derive} continues to predict Adam, Lion, etc. well. (\Apx{app:steplaw}).

\begin{table}[h]
\centering
\footnotesize
\caption{The saving of temporal order does not depend on how the update is made. Each block changes one part of the update and keeps the rest as in the main experiments (LoRA of rank 16, Adam, learning rate $10^{-5}$). Cells read as in \Tbl{tab:steps}: median steps, chronological / shuffled, and the share of steps temporal order saves on time-varying facts.}
\label{tab:robust}
\scriptsize
\setlength{\tabcolsep}{3.5pt}
\renewcommand{\arraystretch}{1.0}
\begin{tabular}{llcccc}
\toprule
 & & \multicolumn{2}{c}{Overwrite} & \multicolumn{2}{c}{Inject} \\
\cmidrule(lr){3-4}\cmidrule(lr){5-6}
Varied & Setting & steps & saved & steps & saved \\
\midrule
\multirow{4}{*}{Optimizer} & Adam & 47.0 / 92.5 & \textbf{$+$49\%} & 32.0 / 54.0 & \textbf{$+$41\%} \\
 & Lion & 51.0 / 108 & \textbf{$+$53\%} & 32.5 / 58.5 & \textbf{$+$44\%} \\
 & SGD & 23.5 / 214 & \textbf{$+$89\%} & 18.0 / 118 & \textbf{$+$85\%} \\
 & SGD + momentum & 7.5 / 33.5 & \textbf{$+$78\%} & 6.0 / 21.5 & \textbf{$+$72\%} \\
\midrule
\multirow{8}{*}{Learning rate} & Adam, $3\!\times\!10^{-5}$ & 18.0 / 32.5 & \textbf{$+$45\%} & 10.0 / 20.0 & \textbf{$+$50\%} \\
 & Adam, $3\!\times\!10^{-6}$ & 148 / 266 & \textbf{$+$45\%} & 87.0 / 169 & \textbf{$+$49\%} \\
 & Adam, $10^{-6}$ & 418 / 781 & \textbf{$+$46\%} & 248 / 492 & \textbf{$+$50\%} \\
 & Lion, $3\!\times\!10^{-6}$ & 168 / 357 & \textbf{$+$53\%} & 106 / 194 & \textbf{$+$45\%} \\
 & Lion, $10^{-6}$ & 497 / 1061 & \textbf{$+$53\%} & 318 / 580 & \textbf{$+$45\%} \\
 & SGD, ${10^{-2}}$ & 3.0 / 22.0 & \textbf{$+$86\%} & 2.0 / 12.5 & \textbf{$+$84\%} \\
 & SGD, $3\!\times\!10^{-3}$ & 8.0 / 71.5 & \textbf{$+$89\%} & 6.0 / 40.0 & \textbf{$+$85\%} \\
 & SGD, $3\!\times\!10^{-4}$ & 76.5 / 708 & \textbf{$+$89\%} & 57.5 / 394 & \textbf{$+$85\%} \\
\midrule
\multirow{3}{*}{LoRA rank} & $r=16$ & 49.5 / 87.0 & \textbf{$+$43\%} & 28.0 / 54.5 & \textbf{$+$49\%} \\
 & $r=4$ & 113 / 184 & \textbf{$+$38\%} & 72.0 / 125 & \textbf{$+$42\%} \\
 & $r=1$ & 234 / 368 & \textbf{$+$36\%} & 152 / 258 & \textbf{$+$41\%} \\
\midrule
\multirow{2}{*}{Full fine-tuning} & Adam, $3\!\times\!10^{-6}$ & 4.0 / 9.0 & \textbf{$+$56\%} & 2.0 / 4.0 & \textbf{$+$50\%} \\
 & Adam, $10^{-6}$ & 10.0 / 24.0 & \textbf{$+$58\%} & 5.0 / 10.5 & \textbf{$+$52\%} \\
\bottomrule
\end{tabular}

\end{table}

\subsection{Derivation of the Step Law}
\label{app:derivation}

This appendix derives \Equ{eq:steplaw} step by step (\Apx{app:steplaw}), checks each approximation (\Apx{app:approx}), and derives the adapter-gradient factorization used in \Fig{fig:cost}c (\Apx{app:factor}).

\subsubsection{From the Trajectory to the Law}
\label{app:steplaw}

\paragraph{Setting.}
For each update, we fix the prompt, the new value, and the competitor, defined as the strongest value the model predicts for that fact before the update (\Sect{sec:setup}).
The adapter parameters $\theta$ start from $\theta_0$, with a fresh optimizer state and no weight decay.
We define
\begin{equation}
L(\theta)=-\log \Pnew(\theta),\quad
C(\theta)=\log \Pnew(\theta)-\log \Pold(\theta),\quad
\Dist=-C(\theta_0)>0,\quad
\gz=\nabla L(\theta_0),
\label{eq:objects}
\end{equation}
where $L$ is the update loss, $C$ is the log-probability margin of the new value over the competitor, $\Dist$ is the initial gap that the update must close, and $\gz$ is the gradient before the first update step.

The update succeeds once the new value becomes more likely than the competitor:
\[
\Nsteps=\min\{k:C(\theta_k)>0\}.
\]
We write $\Delta\theta_k:=\theta_{k+1}-\theta_k$ for the parameter change made by Adam at step $k$.
All probabilities are computed over the full target value.
In practice, we minimize the mean loss over its tokens.

\paragraph{The margin travelled is a path integral.}
Let $\theta(t)$, $t\in[0,T]$, denote the parameter path, with $\theta(k)=\theta_k$ at each integer step, and let $\dot{\theta}(t)$ be its velocity.
By the chain rule, the total change in the margin from the start to step $T$ is
\begin{equation}
C(\theta_T)-C(\theta_0)
=
\int_0^{T}\frac{\mathrm{d}\,C(\theta(t))}{\mathrm{d}t}\,\mathrm{d}t
=
\int_0^{T}
\nabla_{\theta}C\big(\theta(t)\big)^{\!\top}
\dot{\theta}(t)\,\mathrm{d}t .
\label{eq:pathint}
\end{equation}

The integrand has two parts.
The first is the margin gradient,
$\nabla_{\theta}C
=
\nabla_{\theta}\log\Pnew
-
\nabla_{\theta}\log\Pold$,
which we can compute at any parameter value, including $\theta_0$ before the update.
The second is the parameter velocity $\dot{\theta}(t)$.
Unlike the gradient, this velocity has no closed form because Adam generates each iterate from the history of previous gradients through its running moments.

The five steps below simplify this expression until the update cost can be written using quantities available before the update.
In the following, we write $\nabla$ for $\nabla_{\theta}$.

\paragraph{Step 1: break the path into optimizer steps.}
The optimizer gives us the discrete iterates $\theta_0,\theta_1,\ldots$, so we first split the path at these points.
We connect each pair of consecutive iterates with a straight segment:
\begin{equation}
\theta(t)=\theta_k+(t-k)\,\Delta\theta_k,\qquad \dot\theta(t)=\Delta\theta_k,\qquad k\le t\le k+1 .
\label{eq:segment}
\end{equation}
With the substitution $s=t-k$, every segment maps onto the unit interval:
\begin{equation}
C(\theta_T)-C(\theta_0)=\sum_{k=0}^{T-1}\int_0^1 \nabla C(\theta_k+s\,\Delta\theta_k)^{\!\top}\Delta\theta_k\,\mathrm{d}s .
\label{eq:path}
\end{equation}
Because $\nabla C$ is a gradient field, the integral between two iterates depends only on their endpoints.
Thus, the $k$-th term is exactly $C(\theta_{k+1})-C(\theta_k)$, regardless of the path used to connect them.

\paragraph{Step 2: linearize the margin within each step.}
We next simplify what happens between two consecutive iterates.
A segment is short, $\eta$ per coordinate under Step~4, so we approximate the margin gradient along the segment by its value at the start, $\nabla C(\theta_k)$.
This is the first-order Taylor approximation of $C$ around $\theta_k$.
The integral then becomes
\begin{equation}
\Delta C_k:=C(\theta_{k+1})-C(\theta_k)\approx v_k:=\nabla C(\theta_k)^{\!\top}\Delta\theta_k ,
\qquad\text{so}\qquad
C(\theta_T)\approx-\Dist+\sum_{k<T}v_k .
\label{eq:frozen}
\end{equation}
We call $v_k$ the update speed at step $k$: it is how much that step closes the margin.
The update starts with a gap $\Dist$ and succeeds once the accumulated gain reaches this gap, i.e., when $\sum_{k<\Nsteps}v_k\ge\Dist$.
This approximation drops only the curvature of $C$ within each optimizer step.

\paragraph{Step 3: use the first-step speed to estimate the average speed.}
The expression above still requires every $v_k$ along the update trajectory.
To predict the update cost before running the update, we instead use $v_0$, which is available at the start.
The speed does not remain exactly equal to $v_0$ (for example, momentum and the gradient change as $\Pnew$ increases).
We therefore assume that the average speed up to success is a fixed multiple $A$ of the first-step speed,
\begin{equation}
\frac1{\Nsteps}\sum_{k<\Nsteps}v_k\approx A\,v_0
\qquad\Longrightarrow\qquad
\widehat{\Nsteps}=\frac{\Dist}{A\,v_0},\qquad v_0>0 .
\label{eq:average}
\end{equation}
This introduces a fitted parameter $A$.
\Apx{app:approx} measures how well $A v_0$ approximates the average speed.
The remaining two steps express $v_0$ using the initial loss gradient $\gz$.

\paragraph{Step 4: Adam's first update follows the gradient sign.}
We now write the first parameter update explicitly.
At iterate $k$, Adam \citep{kingma2015adam} reads the gradient $g_k=\nabla L(\theta_k)$, updates its two running moments, and applies
\begin{align}
\mu_k&=\beta_1\mu_{k-1}+(1-\beta_1)\,g_k,\qquad
\nu_k=\beta_2\nu_{k-1}+(1-\beta_2)\,g_k^{\odot2},
\label{eq:moments}\\
\Delta\theta_k&=-\eta\,\frac{\hat\mu_k}{\sqrt{\hat\nu_k}+\epsilon},\qquad
\hat\mu_k=\frac{\mu_k}{1-\beta_1^{\,k+1}},\quad
\hat\nu_k=\frac{\nu_k}{1-\beta_2^{\,k+1}},
\label{eq:adam}
\end{align}
with $\mu_{-1}=\nu_{-1}=0$.
At the first step, the optimizer state is still zero.
Thus, $\mu_0=(1-\beta_1)\gz$ and $\nu_0=(1-\beta_2)\gz^{\odot2}$, and the bias corrections give $\hat\mu_0=\gz$ and $\hat\nu_0=\gz^{\odot2}$.
The first update is therefore
\begin{equation}
\Delta\theta_0=-\eta\,\frac{\gz}{|\gz|+\epsilon}\;\approx\;-\eta\,\operatorname{sign}(\gz),
\qquad
v_0=\nabla C(\theta_0)^{\!\top}\Delta\theta_0\approx-\eta\,\nabla C(\theta_0)^{\!\top}\operatorname{sign}(\gz).
\label{eq:sign}
\end{equation}
Ignoring $\epsilon$, each coordinate moves by approximately $\eta$, and the gradient determines its direction rather than its magnitude.

\paragraph{Step 5: express the first-step speed using the loss gradient.}
The expression for $v_0$ still contains the margin gradient
$\nabla C(\theta_0)=\nabla\log\Pnew-\nabla\log\Pold$.
The optimizer, however, directly computes only $\gz=-\nabla\log\Pnew$.
We therefore rewrite $\nabla C$ in terms of $\gz$.

Let $S=\Pnew+\Pold$ be the total probability assigned to the new value and its competitor, and let $\qold=\Pold/S$ be the competitor's share.
Since $\Pnew/S=e^{C}/(1+e^{C})=\sigma(C)$, we can write $\log\Pnew=\log S+\log\sigma(C)$.
Differentiating, with $\sigma'/\sigma=1-\sigma=\qold$, gives
\begin{equation}
\gz=-\nabla\log S(\theta_0)-\qold\,\nabla C(\theta_0),
\label{eq:split}
\end{equation}
The first term changes the total probability of the two values, while the second changes their relative probability.

Solving \Equ{eq:split} for $\nabla C(\theta_0)$ and substituting it into \Equ{eq:sign}, with $\gz^{\!\top}\operatorname{sign}(\gz)=\|\gz\|_1$, gives
\begin{equation}
v_0\approx\frac{\eta}{\qold}\Big[\|\gz\|_1+\nabla\log S(\theta_0)^{\!\top}\operatorname{sign}(\gz)\Big]
\;\approx\;\frac{\eta\,\|\gz\|_1}{\qold}.
\label{eq:speed}
\end{equation}
The last approximation drops the change in the total probability $S$ along the sign step.
This term is small in our updates, only one to two percent of $\|\gz\|_1$.

\paragraph{Putting the steps together.}
Substituting \Equ{eq:speed} into \Equ{eq:average} gives
\begin{equation}
\widehat{\Nsteps}=\frac{\qold\,\Dist}{\eta A\,\|\gz\|_1},
\label{eq:steplawfull}
\end{equation}
where $\Dist$ is the initial gap and $\|\gz\|_1$ determines the initial update speed.
Since $\qold=(1+e^{-\Dist})^{-1}$ is close to one ($\qold \approx 1$), the main text drops this factor and writes \Equ{eq:steplaw}.
Here, $\|\gz\|_1$ is the $L_1$ norm of the initial gradient of $\log\Pnew$ over all adapter parameters, measured directly for every update.

\paragraph{Fitting the constant.}
We fit $A$ once, as the median ratio of $\qold\Dist/(\eta\|\gz\|_1)$ to the observed steps on half of the updates of the undated C at the three scales, which gives $A=1.27$; every prediction in \Sect{sec:derive} uses the other half.
Fitted separately on each model of \Fig{fig:cost}a, $A$ lies between 1.19 and 1.41 for the chronological models and between 1.42 and 1.67 for the shuffled ones, whatever their scale or timestamps.

\subsubsection{How Good Each Approximation Is}
\label{app:approx}

\begin{figure}[t]
\centering
\includegraphics[width=\linewidth]{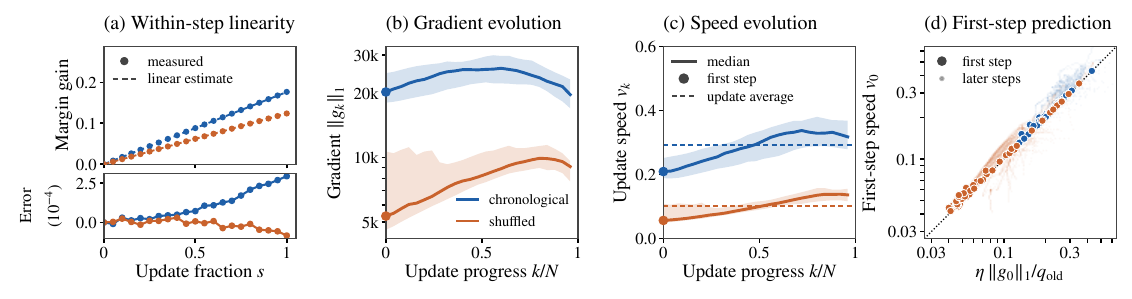}
\caption{The step-law approximations along update trajectories, for time-varying overwrite updates in the 110M C (blue) and S (orange) under Adam. (a) Measured margin gain against its linear estimate along the first update; the lower plot shows their difference in log-probability, in units of $10^{-4}$. Each model is represented by the update with the median first-step error. (b) Gradient norm over update progress $k/N$. (c) Update speed over the same progress: dots mark the first step and dashed lines the median of per-update average speeds. In (b, c), solid curves are medians and shaded bands interquartile ranges. (d) First-step speed against its gradient-based estimate $\eta\|g_0\|_1/\qold$; the diagonal is exact agreement, and faint points show later steps.}
\label{fig:steplaw_checks}
\end{figure}

\Fig{fig:steplaw_checks} shows the approximations at work on the 110M C / S pair.
Step~2 drops the curvature within one update, which is 0.2\% of the margin change at the first step and under 1\% at any later one, so the first-order sum recovers 99.4\% of the margin travelled (\Fig{fig:steplaw_checks} (a)).
Step~3 lets the first speed stand for the average; the average is 1.3 times the first speed in C and 1.8 times in S, whose gradient keeps growing as the update proceeds (\Fig{fig:steplaw_checks} (b) and \Fig{fig:steplaw_checks} (c)).
Steps~4 and~5 replace the first speed by $\eta\|\gz\|_1/\qold$, which matches the measured $v_0$ to within 3\% (\Fig{fig:steplaw_checks} (d)).

\subsubsection{Factorizing the Adapter Gradient}
\label{app:factor}

\Fig{fig:cost} (c) draws the initial gradient module by module.

A LoRA adapter sits on a linear module $m$ with frozen weight $W_m$, and the training sentence passes through the module once per token $p$.
Let $x_{m,p}\in\mathbb{R}^{d_{\mathrm{in}}}$ be the module's input at token $p$ and $y_{m,p}\in\mathbb{R}^{d_{\mathrm{out}}}$ its output.
The adapter adds a low-rank term,
\begin{equation}
y_{m,p}=W_m x_{m,p}+\frac{\alpha}{r}\,B_m A_m x_{m,p}
=W_m x_{m,p}+\frac{\alpha}{r}\,B_m z_{m,p},
\qquad z_{m,p}:=A_m x_{m,p}\in\mathbb{R}^{r},
\label{eq:lora}
\end{equation}
with $A_m\in\mathbb{R}^{r\times d_{\mathrm{in}}}$ drawn at random and $B_m\in\mathbb{R}^{d_{\mathrm{out}}\times r}$ zero at the start, so that the adapter changes nothing until it is trained; $\alpha=r$ in all our updates.
The objective $\log\Pnew$ depends on $B_m$ only through the outputs $y_{m,p}$, so with $\delta_{m,p}:=\partial\log\Pnew/\partial y_{m,p}\in\mathbb{R}^{d_{\mathrm{out}}}$, the backward sensitivity that backpropagation delivers at the module's output, the chain rule over tokens gives
\begin{equation}
\frac{\partial\log\Pnew}{\partial B_m}=\frac{\alpha}{r}\sum_p\delta_{m,p}\,z_{m,p}^{\!\top},
\qquad
\frac{\partial\log\Pnew}{\partial A_m}=\frac{\alpha}{r}\sum_p B_m^{\!\top}\delta_{m,p}\,x_{m,p}^{\!\top} .
\label{eq:gradB}
\end{equation}
The gradient of $A_m$ vanishes at the start of the update because $B_m=0$: the only path from $A_m$ to the output runs through $B_m$, and it carries nothing back.
With $B_m=0$ and $\alpha=r$, the whole gradient the adapter at module $m$ receives is therefore
\begin{equation}
g_m=\sum_p \delta_{m,p}\,z_{m,p}^{\!\top}.
\label{eq:factor}
\end{equation}
The loss gradient $\gz=-\nabla\log\Pnew$ is the adapters' gradients laid end to end, with the $A$-parts all zero; \Fig{fig:cost} (c) draws each adapter's share of it as a cell area, both models on the chronological model's scale.

\section{Details on Four Applications}
\label{app:practical}

\subsection{Mixing Generalists and Specialists}
\label{app:practical-frontier}
\label{app:multi}

\paragraph{Corpora.}
Mix-C in \Sect{sec:frontier} combines a generalist for the history with a specialist for the recent window by splitting each person's records at 2021.
Records before 2021 keep their timestamps and are shuffled.
Records from 2021 onward lose half of their timestamps and keep each person's chronological order.
Mix-S contains the same records but shuffles them all.
Both models are trained at 110M using the recipe in \Apx{app:protocol} and updated as in \Sect{sec:setup}.

\paragraph{Batch and sequential updates.}
A batch update writes many facts into one adapter by training on all of their sentences together.
A sequential update applies a series of such batches, merging each batch into the model before the next.
In \Fig{fig:frontier}(c), a batch contains 2{,}400 facts, while a sequence contains ten batches of fifty.
We measure forgetting after the batch, or after the final batch in the sequence, on facts about other people as in \Apx{app:protocol}.
Across all three settings, Mix-C needs 16--44\% fewer steps than Mix-S and forgets 5--57\% less.

\subsection{Reshaping a Real-World Pre-trained Model}
\label{app:realdata}

\paragraph{Data.}
We continue pre-training Gemma-2-9B-it \citep{team2024gemma} on KairosQA \citep{pilchen2026understanding}, a set of real facts whose answers change over time.
Each fact is a question with a different answer in different years: for example, the head coach of Tottenham Hotspur is Mauricio Pochettino in 2014--2019, José Mourinho in 2020 and Antonio Conte in 2021--2023.
This is like the employer of a person in our biographies.
We use 5,324 such facts whose answer changes between 2001 and 2023, from eight kinds of question (for example, the head coach of a team, the team of a player and the head of a government), and drop the facts about subjects in the CounterFact evaluation set.
Each record asks for one fact in Gemma's chat format, for example ``Who was the head coach of Tottenham Hotspur F.C.\ in 2019? Answer with the name only.''.
An undated record drops the year.

\paragraph{Stage.}
The stage trains on three kinds of text.
The first asks each fact without a year over its recent years, 2016--2023 (52M tokens); this part makes the model a specialist in the latest answer.
The second asks each fact with the year over its whole history, 2001--2023 (35M tokens); this part keeps the model a generalist about the past.
The third is ordinary web text from FineWeb-edu \citep{penedo2024fineweb} (53M tokens), which keeps the model's general ability.
C-C shows the recent years of each fact in time order and mixes in the rest at random; C-S shuffles everything.
The stage runs 5,174 steps, 140M tokens in total, at a constant learning rate of $3\times10^{-6}$.

\paragraph{Updates.}
We adopted 170 \emph{overwrite} updates that are not in the continual pre-training stage.
Each update writes the new answer using the LoRA setup of \Sect{sec:setup} (rank 16, learning rate $10^{-6}$), with either an undated prompt or a dated prompt beginning with ``In 2023''.

\paragraph{Forgetting.}
The replay preserves historical knowledge.
On questions about past years, C-C improves accuracy from 40\% in Base to 99\%.
This shows that the stage strengthens the model's generalist knowledge.
But the replay only protects the facts it covers.
KairosQA facts that never appear in the replay lose about 25\% of the past-year answers that Base originally gets right, in both C-C and C-S, while facts from other sources, such as the CounterFact facts below, do not drop.
This suggests that historical knowledge should be included in the replay if we want the model to keep it.

Finally, the stage causes little loss in general ability (\Tbl{tab:realdata-general}).
On ARC-Easy \citep{clark2018think}, PIQA \citep{bisk2020piqa}, and HellaSwag \citep{zellers2019hellaswag}, C-C is at most 1.1 below Base.
On CounterFact \citep{meng2022rome}, C-C matches Base.

\begin{table}[h]
\centering
\footnotesize
\caption{General ability of Gemma-2-9B-it before (Base) and after the continual pre-training stage (C-C, C-S). Accuracy (\%) on each benchmark and on CounterFact facts, higher is better.
}
\label{tab:realdata-general}
\setlength{\tabcolsep}{5pt}
\begin{tabular}{lcccc}
\toprule
Model & ARC-Easy & PIQA & HellaSwag & CounterFact \\
 & \citep{clark2018think} & \citep{bisk2020piqa} & \citep{zellers2019hellaswag} & \citep{meng2022rome} \\
\midrule
Base & 85.8 & 81.1 & 52.8 & 89.0 \\
C-C & 84.7 & 81.3 & 52.8 & 88.9 \\
C-S & 84.4 & 80.9 & 52.8 & 89.1 \\
\bottomrule
\end{tabular}
\end{table}

\subsection{Improving Published-Knowledge Editors}
\label{app:practical-editing}
\label{app:rome}

\paragraph{Editors.}
We use the official implementations and hyperparameters of ROME \citep{meng2022rome}, MEMIT \citep{meng2023memit}, AlphaEdit \citep{fang2025alphaedit} and ENCORE \citep{gupta2025encore}, with ROME on layer 2 and the others on layers 1--4 of our 12-layer model; AlphaEdit's null-space threshold is raised from 0.02 to 0.1, since at 0.02 layer 4 has no null space.
Each editor makes the same 60 overwrite updates of time-varying facts in Mix-C and Mix-S.
Cost is the FLOPs spent per fact written, where a fact counts as written once its new value's probability passes that of the strongest value the model held before updating.

\paragraph{Results.}
Every editor forgets 50--63\% less on Mix-C than on Mix-S (\Tbl{tab:editors}).

\begin{table}[h]
\centering
\footnotesize
\caption{The four editors on Mix-C and Mix-S, over the same 60 overwrite updates. Each cell is Mix-C / Mix-S, followed by the share Mix-C saves; lower is better. Updating cost: $10^{12}$ FLOPs per fact written. Forgetting, other people: percentage of other people's facts whose answer the update changes. Forgetting, own history: percentage of the updated person's dated history that the update changes. Bold: $p<0.05$.}
\label{tab:editors}
\scriptsize
\setlength{\tabcolsep}{3.5pt}
\begin{tabular}{lcccccc}
\toprule
 & & & \multicolumn{4}{c}{Forgetting (\%)} \\
\cmidrule(lr){4-7}
Editor & Updating cost & saved & Other people & reduced & Own history & reduced \\
\midrule
ROME \citep{meng2022rome} & 5.36 / 5.65 & \textbf{$+$5\%} & 2.1 / 4.2 & \textbf{$+$50\%} & 81.8 / 80.2 & $-$2\% \\
MEMIT \citep{meng2023memit} & 0.78 / 1.44 & \textbf{$+$46\%} & 0.9 / 2.2 & \textbf{$+$59\%} & 14.8 / 22.1 & $+$33\% \\
AlphaEdit \citep{fang2025alphaedit} & 0.77 / 1.31 & \textbf{$+$41\%} & 1.2 / 3.2 & \textbf{$+$63\%} & 16.0 / 24.7 & \textbf{$+$35\%} \\
ENCORE \citep{gupta2025encore} & 0.75 / 1.10 & \textbf{$+$32\%} & 0.8 / 2.1 & \textbf{$+$60\%} & 14.5 / 18.8 & $+$23\% \\
\bottomrule
\end{tabular}

\end{table}

\subsection{Versioned Software Change}
\label{app:toolworld}

\paragraph{Tool-call dataset.}
We build the tool-call dataset in the same way as the biography dataset in \Sect{sec:setup}: a tool corresponds to a person, one of its six fields to an attribute, and the software version to the year.
The dataset contains 25{,}000 tools and 25 versions, from v1 to v25; each tool appears 1{,}000 times, and a fixed set of tools is held out for injection updates.

The six fields follow the same change rates as the biography attributes: package and function never change, host and authentication header change a few times, and path and input parameter change often.
Each record gives the six fields of one tool at one version:
\par\begingroup\small\addtolength{\leftskip}{1.2em}\addtolength{\rightskip}{1.2em}\vspace{3pt}\noindent
In version v8, the Flint Payments API uses package \underline{flintpay\_sdk} and function \underline{revoke\_refund}.
It carries host \underline{api.flintpay.com} and authentication header \underline{X-Api-Key}.
It is called at path \underline{/v1/refunds} with input parameter \underline{refund\_id}.
\par\vspace{3pt}\endgroup

\paragraph{Training and updates.}
For pre-training from scratch, we follow \Sect{sec:frontier}: Mix-C shuffles the history (v1--v21) with version labels, while keeping each tool's recent records (v22--v25) in version order with half of the version labels removed.
Mix-S contains the same records but shuffles them all.

For continual pre-training, we follow \Apx{app:cpt}: starting from a Base model trained on shuffled records with version labels, C-C presents a window of new versions in order and mixes them with a shuffled replay of the history, while C-S shuffles the same records.
Every model receives single updates both with and without the version in the prompt.

\paragraph{Results.}
We find that temporal order brings the same benefits to tool calls as to biographies.
All models keep the old versions well: with the version in the prompt, query accuracy stays between 0.999 and 1.00.
Without the version, Mix-C reaches 0.98 accuracy on the latest version, compared with 0.26 for Mix-S.
After continual pre-training, C-C reaches 0.99, compared with 0.76 for C-S.
The chronological models also update faster (\Tbl{tab:toolworld}).
Without the version in the prompt, they need 14--48\% fewer steps to update a changing field.
With the version in the prompt, the saving becomes smaller, at 7--29\%.

\begin{table}[h]
\centering
\footnotesize
\caption{Median steps to update a field in the tool-call world, for slow and fast fields. Rows: pre-training from scratch (Mix-C / Mix-S) and continual pre-training (C-C / C-S). \emph{saved} is the share of steps the chronological model saves; bold: $p<0.05$.}
\label{tab:toolworld}
\scriptsize
\setlength{\tabcolsep}{3.5pt}
\begin{tabular}{lcccccccc}
\toprule
 & \multicolumn{4}{c}{No version in prompt} & \multicolumn{4}{c}{Version in prompt} \\
\cmidrule(lr){2-5}\cmidrule(lr){6-9}
 & \multicolumn{2}{c}{Overwrite} & \multicolumn{2}{c}{Inject} & \multicolumn{2}{c}{Overwrite} & \multicolumn{2}{c}{Inject} \\
\cmidrule(lr){2-3}\cmidrule(lr){4-5}\cmidrule(lr){6-7}\cmidrule(lr){8-9}
Fact & steps & saved & steps & saved & steps & saved & steps & saved \\
\midrule
\multicolumn{9}{l}{\textit{Pre-training from scratch: Mix-C / Mix-S}} \\
\quad slow & 53.0 / 102 & \textbf{$+$48\%} & 38.5 / 64.0 & \textbf{$+$40\%} & 55.0 / 77.0 & \textbf{$+$29\%} & 40.5 / 57.0 & \textbf{$+$29\%} \\
\quad fast & 47.0 / 84.0 & \textbf{$+$44\%} & 33.5 / 47.5 & \textbf{$+$29\%} & 48.5 / 68.5 & \textbf{$+$29\%} & 32.0 / 42.0 & \textbf{$+$24\%} \\
\midrule
\multicolumn{9}{l}{\textit{Continual pre-training: C-C / C-S}} \\
\quad slow & 49.5 / 70.5 & \textbf{$+$30\%} & 37.5 / 45.0 & \textbf{$+$17\%} & 55.5 / 61.5 & \textbf{$+$10\%} & 40.0 / 46.5 & \textbf{$+$14\%} \\
\quad fast & 43.5 / 64.5 & \textbf{$+$33\%} & 34.5 / 40.0 & \textbf{$+$14\%} & 46.0 / 49.5 & \textbf{$+$7\%} & 32.0 / 37.0 & \textbf{$+$14\%} \\
\bottomrule
\end{tabular}

\end{table}

\subsection{Reshaping a Pre-trained Model in Synthetic world}
\label{app:cpt}

The continual pre-training stage of \Sect{sec:cpt}, applied to the 110M SD\half{} model of \Sect{sec:shape}; \Sect{sec:apply} follows the same stage.

\begin{figure}[t]
\centering
\includegraphics[width=0.62\linewidth]{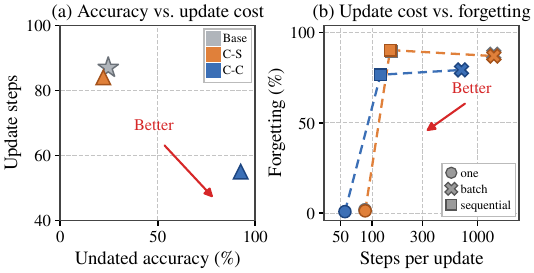}
\caption{Reshaping the 110M SD\half{} model. (a) Undated query accuracy against the median update steps for Base, C-S and C-C. (b) Steps per update against forgetting for single, batch and sequential updates of Base (grey), C-S (orange) and C-C (blue); lower left is better.}
\label{fig:reshape-110m}\label{fig:reshape-forgetting}
\end{figure}

\paragraph{Dataset setup.}
Base is the 110M SD\half{} model of \Sect{sec:shape}.
It is pre-trained on shuffled records, and only a random half of them carry a timestamp.
Continual pre-training shows Base only records it has already seen, in two parts.
The historical part has 100 timestamped records per person in shuffled order, and states every value the person has held.
The recent part has 200 undated records per person in chronological order, and includes every change of value.
C-C mixes the two parts and keeps the order within each person; C-S holds exactly the same records, shuffled.
Each stage is one pass of about 2800 steps at batch 2{,}560 (about 0.6B tokens) at a constant learning rate of $6\times10^{-5}$, the final rate of pre-training.

\paragraph{Results.}
We find that a short chronological stage gives Base a specialist for the present while keeping its past.
C-C raises Base's undated query accuracy from 25\% to 93\% and keeps its dated accuracy at 98\%.
C-S, trained on the same records, reaches only 22\% (\Fig{fig:reshape-110m} (a)).
C-C also updates faster: it needs up to 1.6$\times$ fewer steps than Base for single updates, $2.1\times$ fewer in a batch and $1.3\times$ fewer in a sequence.
It also forgets 10--15\% less (\Fig{fig:reshape-110m} (b)).

\end{document}